\documentclass[final,5p,times]{elsarticle}
 
\usepackage{amsmath,amssymb}
\usepackage{booktabs}
\usepackage{makecell}

\usepackage{multirow}
\usepackage{graphicx}
\usepackage[hidelinks]{hyperref}
\usepackage{xurl}   %
\usepackage{microtype}  %
\usepackage{float}      %
\usepackage{placeins}   %
\usepackage{capt-of}    %
\makeatletter
\newenvironment{pairedfloat}[1]{\par\def\@captype{#1}}{\par}
\makeatother
\usepackage{array}      %
\providecommand{\bibinfo}[2]{#2}
\newcommand{\suppdoi}[1]{doi:\path{#1}}
\newcounter{suppref}
\renewcommand{\thesuppref}{S\arabic{suppref}}
\newenvironment{suppreferences}%
  {\section*{Supplementary references}%
   \begin{list}{[\thesuppref]}%
     {\usecounter{suppref}\setlength{\leftmargin}{2.2em}%
      \setlength{\labelwidth}{1.8em}\setlength{\labelsep}{0.4em}%
      \setlength{\itemsep}{2pt}\setlength{\parsep}{0pt}}\small}%
  {\end{list}}

\journal{Medical Image Analysis}

\begin{document}

\begin{frontmatter}

\title{Accurate MRI-free PET Quantification and Reliability Assessment via CT-to-MRI Synthesis and Conformal Prediction}

\author[snu]{Jungwoo Seo}
\ead{jungwoo.seo95@gmail.com}
\author[snu]{Seung Yun Choi}
\author[chosun]{Kun Ho Lee}
\author[yonsei]{Seun Jeon\corref{cor1}}
\ead{seun.jeon@gmail.com}
\author[yonsei]{Byoung Seok Ye\corref{cor1}}
\ead{ROMEL79@yuhs.ac}
\author[snu]{Jiook Cha\corref{cor1}}
\ead{connectome@snu.ac.kr}
\cortext[cor1]{Corresponding authors.}

\address[snu]{Seoul National University, Seoul, Republic of Korea}
\address[chosun]{Chosun University, Gwangju, Republic of Korea}
\address[yonsei]{Yonsei University, Seoul, Republic of Korea}

\begin{abstract}
Standardized uptake value ratio (SUVR) quantification of amyloid PET commonly
relies on a subject-specific T1-weighted MRI, yet a usable MRI is not always
available. We therefore combined CT-to-MRI synthesis with conformal prediction for MRI-free PET quantification with reliability assessment. A slice-consistent Brownian Bridge Diffusion Model was strengthened with multi-dataset training and data-source conditioning to enrich the anatomical prior learned from MRI, and with expanded multi-slice CT context to exploit the through-plane continuity of subject-specific anatomy in CT. Synthetic MRI was used for FreeSurfer-based anatomical parcellation
and PET quantification, while deep-ensemble uncertainty was calibrated with
conformal prediction to estimate a plausible range for the corresponding real MRI-based SUVR.
On $^{11}$C-PiB PET from OASIS-3, the proposed method achieved the highest
synthesis and parcellation fidelity among the compared methods. Mean cortical SUVR from the proposed synthetic MRI agreed closely with the real MRI-based reference
(MAE $0.016 \pm 0.015$, Pearson $r=0.9987$) and yielded the same amyloid-positivity classification as the real MRI in all 87 test subjects at the prespecified cut-off. Conformal prediction achieved empirical coverage close to the prespecified levels, and at 90\% target coverage, 85 of 87 subjects had prediction intervals
that supported an unambiguous amyloid-positivity decision. Zero-shot evaluation on an independent external cohort showed that the proposed CT-to-MRI model could support PET quantification without retraining, although conformal calibration required target-cohort recalibration to recover the
intended coverage. These findings support CT-to-MRI-based MRI-free PET quantification as a promising approach for providing accurate SUVR measurements and an explicit assessment of their reliability.
\end{abstract}

\begin{keyword}
amyloid PET \sep MRI-free PET quantification \sep CT-to-MRI synthesis \sep
diffusion models \sep conformal prediction \sep uncertainty quantification
\end{keyword}

\end{frontmatter}

\section{Introduction}

Amyloid-$\beta$ deposition is one of the earliest detectable
pathological changes in Alzheimer's disease and may begin years before
measurable cognitive decline \citep{bateman2012clinical,jack2024revised}. Positron emission tomography (PET)
enables this molecular pathology to be visualized and quantified in
vivo \citep{klunk2004imaging}. As Alzheimer's disease is increasingly
defined and staged according to its underlying biological pathology,
biomarker-based assessment has become increasingly important
\citep{jack2024revised}. Quantitative amyloid PET can characterize
overall pathological burden, regional deposition patterns relevant to
disease staging \citep{grothe2017invivo}, and longitudinal changes in
amyloid accumulation \citep{villain2012regional}.

Amyloid PET is commonly quantified using the standardized uptake value
ratio (SUVR), which normalizes tracer uptake in cortical target regions
to uptake in a relatively amyloid-sparse reference region. Accurate
SUVR estimation requires the target and reference regions to be defined
according to each subject's anatomy. Because PET has limited spatial
resolution and weak anatomical contrast, conventional quantification
pipelines typically rely on a subject-specific T1-weighted MRI for
anatomical parcellation
\citep{lopresti2005simplified,su2013quantitative}.

A usable MRI, however, is not always available because of implanted
devices, claustrophobia, patient motion, acquisition burden, or
additional cost \citep{nguyen2020prevalence}. This has motivated
MRI-free PET quantification methods based on PET spatial
normalization \citep{edison2013,landau2023quantification,
kang2023fast}, low-dose attenuation CT \citep{
presotto2018ct,matsuda2021amyloid,choo2025deep}, or CT-to-MRI synthesis
\citep{kim2025automated}.
PET-only methods provide limited subject-specific anatomy, whereas CT
offers insufficient soft-tissue contrast for fine cortical
parcellation. CT-to-MRI synthesis is therefore particularly promising:
it can complement the anatomical evidence available in CT with
structural knowledge learned from MRI while retaining established
MRI-based segmentation and quantification pipelines.

Despite this promise, CT-to-MRI synthesis remains intrinsically ill-posed
because CT provides substantially less information than T1-weighted MRI about
gray--white matter boundaries and cortical folding. Multiple anatomically
plausible MR images may therefore be consistent with the same CT observation,
making faithful recovery of subject-specific fine anatomy inherently
challenging. More generally, medical image translation models can produce
visually plausible outputs without necessarily preserving all
subject-specific anatomical features
\citep{cohen2018distribution}. This ambiguity is particularly important for
MRI-free PET quantification because anatomical errors in the synthesized MRI
can propagate through brain parcellation and ultimately affect regional SUVR
estimates.

Conceptually, CT-to-MRI synthesis can be viewed as estimating the conditional
distribution of possible MR images given the observed CT:

\[
p(\mathbf{MRI}\mid\mathbf{CT})
\propto
p(\mathbf{CT}\mid\mathbf{MRI})p(\mathbf{MRI}).
\]

Here, $p(\mathbf{MRI})$ represents anatomical prior knowledge learned from MRI
data, whereas $p(\mathbf{CT}\mid\mathbf{MRI})$ reflects compatibility with the
subject-specific anatomical information available in CT. Faithful synthesis
therefore requires both a sufficiently rich anatomical prior and sufficient
subject-specific evidence from the input CT to identify the anatomy most
consistent with the individual subject.

Previous work demonstrated that CT-derived synthetic MRI can be incorporated
into an MRI-based PET quantification pipeline, establishing the feasibility of
CT-to-MRI-based MRI-free PET quantification \citep{kim2025automated}.
Separately, slice-consistent Brownian Bridge diffusion improved CT-to-MRI
synthesis by jointly modeling adjacent slices and promoting through-plane
consistency \citep{choo2024slice,li2023bbdm}. Nevertheless, the anatomical
prior learned from limited training data and the subject-specific information
extracted from a narrow CT input may still be insufficient for recovering
complex individual cortical anatomy.

In this study, we strengthen both sources of information. We use
multi-dataset CT--MRI training to learn from a broader distribution of brain
anatomy and introduce data-source conditioning so that this broader prior can be learned without conflating dataset-specific image characteristics. We also provide a wider multi-slice CT context to make
greater use of subject-specific anatomical information available across
adjacent slices. We hypothesize that a richer anatomical prior together with
stronger subject-specific CT constraints will produce more faithful synthetic
MRI and, in turn, improve anatomical parcellation and MRI-free PET
quantification.

Existing MRI-free PET quantification methods, including CT-to-MRI-based quantification \citep{kim2025automated}, are typically evaluated by their overall agreement with the real MRI-based reference but do not provide, for each subject, a plausible range for the SUVR that would have been obtained with that subject's real MRI.
High overall agreement, however, does not
indicate how reliable an MRI-free SUVR estimate is for an individual subject.
In the intended MRI-free setting, the real MRI is unavailable, so the
discrepancy between an MRI-free SUVR estimate and the corresponding real
MRI-based value cannot be directly measured. This is particularly important
near an amyloid-positivity decision cut-off, where even a small quantification
error may affect the resulting classification.

To address this limitation, we combine deep-ensemble uncertainty with split
conformal prediction
\citep{lakshminarayanan2017simple,vovk2005algorithmic,
lei2018distribution,angelopoulos2023gentle}. Model-derived uncertainty has
been used to characterize output reliability in medical image segmentation
\citep{mehrtash2020confidence} and neuroimage enhancement
\citep{tanno2021uncertainty}, while conformal prediction has been applied to
assess the reliability of diagnostic predictions in AI-assisted pathology
\citep{olsson2022estimating}. In our setting, these ideas are used to assess
the reliability of CT-to-MRI-based PET quantification by estimating a
plausible range for the corresponding real MRI-based SUVR. Comparing this range with the amyloid-positivity decision cut-off
allows cases to be identified as confidently negative, confidently positive,
or indeterminate, providing a way to assess decision reliability when the real
MRI is unavailable.

For clinical applicability, the proposed approach should remain reliable
beyond the datasets used for model development. We therefore evaluated
zero-shot generalization on an independent external cohort and examined
whether the conformal calibration could also be transferred to this new
setting.

Accordingly, this study makes four main contributions:

\begin{enumerate}
    \item \textbf{Improved CT-to-MRI synthesis.}
    We strengthen slice-consistent CT-to-MRI synthesis by enriching the anatomical prior learned from MRI through multi-dataset training with data-source conditioning, and by exploiting the through-plane continuity of subject-specific anatomy in CT through an expanded multi-slice CT context.

    \item \textbf{More accurate MRI-free PET quantification.}
    We show that the improved anatomical inference yields more faithful
    anatomical parcellation and closer agreement with real MRI-based composite and regional SUVR quantification than PET-only and CT-based baselines and existing CT-to-MRI-based approaches.

    \item \textbf{Reliability assessment using conformal prediction.}
    We construct calibrated prediction intervals for composite and regional
    SUVR estimates and use these intervals to identify cases in which the
    reliability of an MRI-free amyloid-positivity decision is uncertain.

    \item \textbf{Zero-shot external validation.} We evaluate CT-to-MRI synthesis, PET quantification, and conformal reliability assessment on an independent external cohort never used for training, demonstrating the feasibility of using the synthesis model for PET quantification without retraining and showing that conformal calibration requires target-cohort recalibration.
\end{enumerate}

 The complete approach is evaluated using $^{11}$C-PiB amyloid PET from OASIS-3 \citep{lamontagne2019oasis3} and, zero-shot, using $^{18}$F-florbetaben PET from the independent Gwangju Alzheimer's \& Related Dementias (GARD) cohort \citep{choi2026gwangju}, with PET quantification based on each subject's real MRI serving as the reference in both cohorts.

\begin{figure*}[t]
\centering
\includegraphics[width=\linewidth]{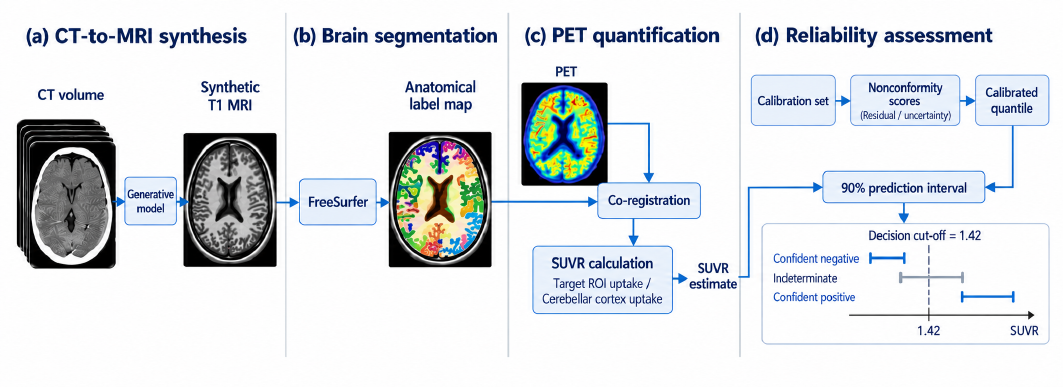}
\caption{
Overview of the MRI-free PET quantification and reliability assessment
framework. The framework consists of four stages: CT-to-MRI synthesis,
anatomical parcellation of the synthesized MRI, SUVR quantification, and
conformal prediction for reliability-aware decision assessment with respect to
the prespecified decision cut-off. This schematic was generated with OpenAI's gpt-image model (version~2.0) based on the authors' design and checked by the authors. The brain images are illustrative and do not depict study data.
}
\label{fig:overview}
\end{figure*}

\section{Materials and Methods}

\subsection{Overview}
\label{sec:methods:overview}

The proposed pipeline replaces the unavailable T1-weighted MRI with a synthesized MRI from a brain CT and then applies the same PET
quantification procedure used in the conventional MRI-based pipeline
(Fig.~\ref{fig:overview}). The framework
consists of four stages. First, a CT-to-MRI synthesis model generates a
subject-specific T1-weighted MRI from the input CT. Second, the synthetic MRI
is processed with FreeSurfer \texttt{recon-all}
\citep{fischl2012freesurfer} to obtain subject-specific anatomical
parcellations. Third, amyloid PET uptake is quantified as standardized uptake
value ratios (SUVRs) using subject-specific anatomical regions and the
cerebellar cortex as the reference region \citep{su2013quantitative}.
Fourth, uncertainty in the MRI-free SUVR estimate is quantified using a deep
ensemble \citep{lakshminarayanan2017simple} and calibrated with conformal
prediction \citep{angelopoulos2023gentle} to construct a prediction interval
for the corresponding real MRI-based SUVR. The interval is then used to
distinguish confident from indeterminate amyloid-positivity decisions.
The downstream parcellation and PET quantification procedures are identical
to those used with the real MRI, with only the anatomical image replaced.

Building on the slice-consistent BBDM framework
\citep{choo2024slice,li2023bbdm}, we introduce multi-dataset training,
data-source conditioning, and expanded multi-slice CT context to better
capture anatomical variation and subject-specific information from CT.
The resulting CT-to-MRI-based PET quantification pipeline is evaluated
zero-shot on a fully independent external cohort without retraining or
adaptation of the synthesis model. We additionally examine the transferability
of conformal calibration to the external cohort and the effect of
target-cohort recalibration on external coverage.

\subsection{Datasets}
\label{sec:methods:data}

To develop a CT-to-MRI generative model that captures diverse anatomical
and acquisition characteristics and to evaluate its generalization to a
cohort independent of the datasets used for model training, we used two publicly available datasets, one in-house
clinical cohort, and one independent external cohort. OASIS-3
\citep{lamontagne2019oasis3} served as the primary dataset for CT-to-MRI
model development and evaluation of the CT-to-MRI-based MRI-free PET
quantification pipeline. SynthRAD2023 \citep{thummerer2023synthrad} provided
additional paired CT--MRI data for synthesis training. The internal Severance Hospital cohort was used for
CT-to-MRI synthesis training and evaluation, providing an additional
clinically acquired CT--MRI dataset with potential applicability to
downstream PET quantification. 
The Gwangju
Alzheimer's \& Related Dementias (GARD) cohort \citep{choi2026gwangju} was
reserved as a fully independent external cohort to evaluate zero-shot
generalization to an unseen cohort without retraining or fine-tuning.

\paragraph{OASIS-3}
OASIS-3 is a publicly available longitudinal multimodal neuroimaging dataset
spanning normal aging and Alzheimer's disease
\citep{lamontagne2019oasis3}. We used low-dose CT acquired as part of PET/CT
examinations together with T1-weighted MRI for CT-to-MRI model development
and evaluation.

CT--MRI pairs were constructed by matching each CT session to the temporally
closest T1-weighted MRI from the same subject within 365 days, yielding
1,441 candidate pairs from 879 subjects. Subjects were assigned at the
subject level to a training set, a validation set for model selection, a
calibration set for conformal prediction, and a test set for final
evaluation. All available CT--MRI pairs from training subjects were retained,
whereas one pair per subject, selected by the smallest CT--MRI acquisition
interval, was used for the validation, calibration, and test sets. This
resulted in 1,052 CT--MRI pairs before quality control. After excluding
22 pairs because of data or preprocessing abnormalities, including
registration failure, corrupted volumes, or incomplete brain coverage,
1,030 usable pairs remained.

For PET quantification, we used $^{11}$C-PiB PET and selected subjects whose
PiB PET and paired T1-weighted MRI were acquired on the same day. This
yielded 93 subjects for conformal calibration and 87 test subjects for the
primary MRI-free PET quantification analysis.

\paragraph{SynthRAD2023}
SynthRAD2023 is a publicly available multi-institutional dataset developed
for synthetic CT generation in radiotherapy and provides paired CT and MRI
data from multiple contributing centers \citep{thummerer2023synthrad}.
We used the brain subset of Task~1, which provides paired MRI--CT data for
MRI-to-CT synthesis, comprising 180 paired CT--MRI cases.
Three cases were excluded because of CT or preprocessing failures, leaving
177 usable pairs for CT-to-MRI synthesis training.

\paragraph{Severance Hospital cohort}
The Severance Hospital cohort is an in-house clinical dataset comprising
paired CT and T1-weighted MRI acquired at Severance Hospital, Seoul,
Republic of Korea. CT and MRI examinations were paired when acquired within one year. The initial cohort included 230 CT-MRI pairs. After excluding 18
pairs because of incomplete brain coverage, image abnormalities, or
preprocessing failure, 212 usable pairs remained. CT images were acquired using a GE Discovery 600 PET/CT scanner (GE Healthcare, Milwaukee, WI, USA) with the following parameters: tube voltage, 120 kVp; tube current, 200 mA; gantry rotation time, 0.5 s; total collimation width, 10 mm; reconstruction field of view, 250 mm; and reconstructed matrix, 512 × 512 with 151 slices; voxel size, 0.488 × 0.488 × 1 mm³. T1-weighted MRI was acquired using a Philips 3.0-T MRI scanner (Philips Achieva; Philips Medical Systems, Best, The Netherlands) with a SENSE head coil (SENSE factor = 2) and a three-dimensional T1-weighted turbo-field echo sequence with the following parameters: axial acquisition matrix, 224 × 224; reconstructed matrix, 256 × 256 with 170 slices; voxel size, 0.859 × 0.859 × 1 mm³; field of view, 220 mm; echo time, 4.6 ms; repetition time, 9.8 ms; and flip angle, 8°.

\paragraph{GARD cohort}
The Gwangju Alzheimer's \& Related Dementias (GARD) cohort is a longitudinal,
community-based South Korean cohort established for multimodal investigation
of aging and Alzheimer's disease \citep{choi2026gwangju}. GARD served as a fully
independent external evaluation cohort and was not used for CT-to-MRI model
training, validation, or model selection, or for the primary conformal
calibration performed using OASIS-3.

For the present study, we identified subjects with
$^{18}$F-florbetaben amyloid PET/CT and T1-weighted MRI acquired within
90 days. The CT was obtained from the same PET/CT examination as the amyloid
PET, and the temporally closest T1-weighted MRI was paired with each PET/CT
examination. For subjects with multiple eligible examinations, the pair with
the smallest PET/CT--MRI acquisition interval was retained, yielding one
CT--MRI--PET set per subject and a final external cohort of 84 subjects.

PET/CT examinations were acquired at two centers using GE Discovery ST and
Discovery STE systems. CT acquisition characteristics differed between the
two sites, with predominant slice thicknesses of 3.75 and 2.5~mm and in-plane
pixel spacings of approximately 0.98 and 0.49~mm, respectively. All
T1-weighted MRI scans were acquired on a 3-T Siemens Skyra system at Chosun
University Hospital. The trained CT-to-MRI models were applied to GARD without
retraining or adaptation.

\paragraph{Data splits}
All data splits were defined at the subject level to prevent information leakage. OASIS-3 was divided into training (773 pairs from 620 subjects), validation (20 subjects), calibration (120 subjects), and test (117 subjects) sets, with multiple longitudinal pairs from the same subject retained only in the training set. The Severance Hospital cohort was divided into training (180), validation (10), and test (22) sets, and all 177 SynthRAD2023 pairs were used for training.

For CT-to-MRI synthesis, the model was trained on the resulting 1,130 CT--MRI pairs, model selection used the 30 validation pairs, and synthesis performance was evaluated on the OASIS-3 and Severance test sets.

For PET quantification and reliability assessment, the OASIS-3 calibration and test sets were both restricted to subjects whose PiB PET and paired T1-weighted MRI were acquired on the same day. This yielded 93 subjects for conformal calibration and 87 subjects for the primary MRI-free PET quantification analysis. The independent GARD cohort was used only for external evaluation and was not involved in model training, model selection, or the primary conformal calibration.

\begin{figure*}[t]
\centering
\includegraphics[width=\linewidth]{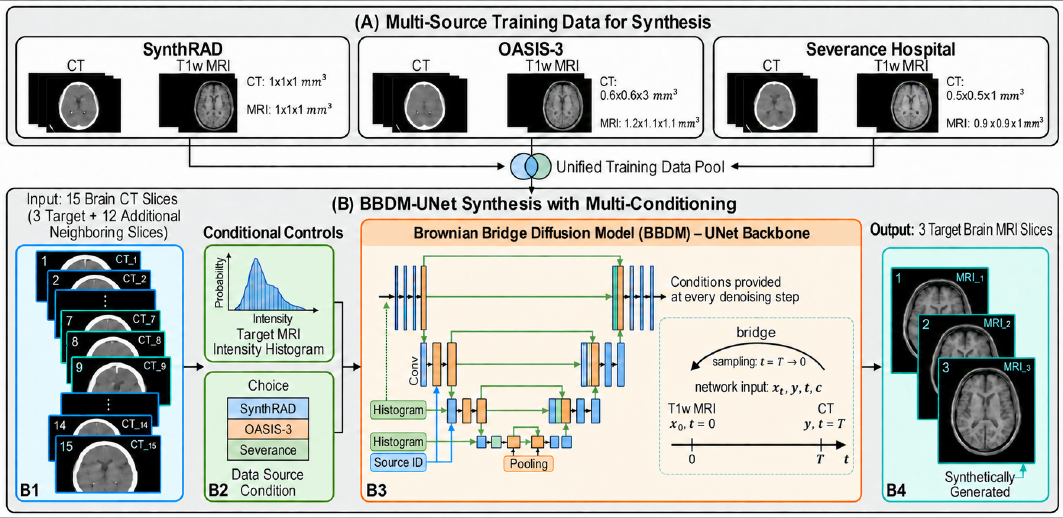}
\caption{Overview of the proposed CT-to-MRI generative model for faithful CT-to-MRI translation, which integrates multi-dataset training, data-source conditioning, and wider anatomical context within a Brownian Bridge Diffusion Model. Paired CT and T1-weighted MRI from three datasets are combined into a unified training pool. (B1) 15 consecutive brain CT slices are used to predict the central three T1-weighted MRI slices, providing broader anatomical context for anatomically faithful synthesis. (B2) Two conditioning signals are provided: the target MRI intensity histogram, which promotes slice-consistent synthesis following Choo et al. (2024) \citep{choo2024slice}, and the data-source condition, which accounts for inter-dataset heterogeneity. (B3) CT-to-MRI synthesis is performed using a Brownian Bridge Diffusion Model with a U-Net backbone, where both conditioning signals are injected into every denoising step. (B4) The model generates three consecutive T1-weighted MRI slices. The segmentation-guided training module explored during model development is omitted because it is not part of the final model. This schematic was generated with OpenAI's gpt-image model (version~2.0) based on the authors' design and checked by the authors. The brain images are illustrative and do not depict study data.}
\label{fig:arch}
\end{figure*}

\subsection{Image preprocessing}
\label{sec:methods:preprocessing}

For model development and retrospective evaluation, paired CT and
T1-weighted MRI volumes were processed in a common space before CT-to-MRI
synthesis. Both modalities were first resampled to
1-mm isotropic resolution. To reduce unnecessary spatial mismatch between
the paired CT and MRI volumes and to enable voxel-wise evaluation against the real MRI, both modalities were then rigidly aligned to
a common MNI152 space. Specifically, the CT was first rigidly registered to
the corresponding T1-weighted MRI using \texttt{elastix}
\citep{klein2010elastix}. A brain mask was then extracted from the
T1-weighted MRI using SynthStrip \citep{hoopes2022synthstrip} and applied to
both the T1-weighted MRI and the registered CT. The T1-weighted MRI was
subsequently rigidly aligned to MNI152 space, and the same transformation
was applied to the paired CT. In a practical MRI-free setting where a
T1-weighted MRI is unavailable, the CT can instead be rigidly aligned
directly to MNI152 space, with a brain mask derived from the CT, before model inference.

Following \citet{kim2025automated}, CT intensities were clipped to the
range of $[-100, +200]$ HU, whereas T1-weighted MRI intensities were clipped
to the 0.01th--99.99th percentile range. Each modality was then min--max
normalized. N4 bias-field correction was not applied in this study,
following the preprocessing conventions of \citep{thummerer2023synthrad} and \citet{kim2025automated}.

\subsection{CT-to-MRI synthesis}
\label{sec:methods:synthesis}

We developed the CT-to-MRI synthesis model based on the slice-consistent
Brownian Bridge Diffusion Model (BBDM) of \citet{choo2024slice}, which
builds on the Brownian Bridge diffusion framework of
\citet{li2023bbdm}. BBDM formulates paired image-to-image translation as a diffusion process
between the source and target modalities. The original framework jointly generates three
consecutive T1-weighted MRI slices from the corresponding three axial CT
slices and uses MRI histogram-based style conditioning together with
ISTA-style sampling across overlapping slice groups. We retained these
components as the base synthesis framework.

Building on this framework, we evaluated four design strategies intended to
compensate for the limited anatomical information available in CT.

\textbf{Multi-dataset training (H1)} jointly used paired CT--MRI data from
OASIS-3, SynthRAD2023, and the Severance Hospital cohort. Training on a
larger and more diverse set of subjects was expected to expose the model to
a broader range of brain anatomy and acquisition characteristics, thereby
strengthening the anatomical prior learned for CT-to-MRI synthesis.

\textbf{Data-source conditioning (H2)} was introduced to account for
heterogeneity across the training datasets. Although multi-dataset training
broadens the available anatomical information, systematic differences in
scanner characteristics, acquisition protocols, and image appearance across
datasets may also interfere with learning a consistent CT-to-MRI mapping.
We therefore provided the source dataset of each training sample as an
additional conditioning variable alongside the MRI histogram condition at
each denoising step, allowing the model to account for dataset-specific
variation during training.

\textbf{Expanded multi-slice anatomical context (H3)} increased the
through-plane CT context available for each prediction. Because gray--white
matter boundaries and complex cortical folding are only weakly represented
within an individual CT slice, we hypothesized that incorporating adjacent
slices would provide complementary subject-specific anatomical information
through the spatial continuity of brain structures across slices.
Accordingly, within a practical computational budget, we evaluated
progressively wider input contexts ranging from 3 to 15 consecutive axial CT
slices. We selected 15 slices for the final model based on validation
performance (Section~\ref{sec:supp:context}).

\textbf{Segmentation-teacher feedback (H4)} was explored as an auxiliary
anatomical regularization strategy. We hypothesized that explicitly
encouraging agreement in gray--white matter and other anatomical boundaries
could promote anatomically faithful synthesis beyond image-level appearance
alone. A frozen pretrained segmentation model was therefore used to provide
an additional anatomical training signal. The teacher was used only during
training and introduced no additional requirement at inference. Details of the segmentation-teacher
model, auxiliary loss formulation, and weight selection are provided in
Supplementary Section~\ref{sec:supp:teacher}.

\textbf{Final model} (Fig.~\ref{fig:arch}) combined multi-dataset training (H1),
data-source conditioning (H2), and the 15-slice CT context (H3), based on the
validation analyses (Supplementary Sections~\ref{sec:supp:ablation-sec}
and~\ref{sec:supp:context}). Segmentation-teacher feedback (H4) was excluded
because it did not provide a significant additional improvement.

\paragraph{Training and inference}
During training, MRI histogram conditioning was obtained from the paired
target T1-weighted MRI, and the source dataset label was additionally
provided for data-source conditioning. The final model used 15 consecutive
axial CT slices to generate the corresponding central three T1-weighted MRI
slices.

At inference, where the target MRI is unavailable, the mean MRI histogram
estimated from the training data was used as the style condition. No subject-specific MRI image or histogram was used as a model input at inference. The source
dataset label corresponding to the input CT was also provided to the model.
For GARD, whose CT acquisition characteristics were most similar to those of
OASIS-3 among the training datasets, the OASIS-3 source label was used for
zero-shot inference. Full three-dimensional synthetic MRI volumes were
reconstructed using the slice-consistent ISTA-style sampling procedure of
\citet{choo2024slice}. The same inference procedure was applied to all test
cohorts, including GARD, without retraining or fine-tuning. Detailed
optimization and implementation settings are provided in the Supplementary
Material (\ref{sec:supp:impl}).

\paragraph{Baseline models}
We compared the proposed model with four CT-to-MRI synthesis baselines
spanning representative image-to-image translation approaches: a
least-squares GAN (LSGAN;~\citealp{mao2017least}), an adaptive latent
diffusion model (ALDM;~\citealp{kim2024adaptive}), the CT-to-MRI synthesis
method developed for MRI-free PET quantification by
\citet{kim2025automated}, and the slice-consistent BBDM of
\citet{choo2024slice}.

Each baseline was trained and evaluated separately on each dataset, following
the conventional single-dataset setting used in the corresponding prior
methods. Thus, models evaluated on OASIS-3 were trained using only the
OASIS-3 training set, whereas those evaluated on Severance were trained
using only the Severance training set. To evaluate the effect of
multi-dataset training (H1), the proposed model was instead trained jointly
on OASIS-3, SynthRAD2023, and Severance. The contribution of H1, together
with the subsequent effects of H2 and H3, was examined in the cumulative
ablation analysis on the validation sets
(Section~\ref{sec:supp:ablation-sec}).

Model-specific training settings followed the corresponding original
publications where applicable, with any modifications or additional tuning
reported in the Supplementary Material (Section~\ref{sec:supp:baselines}).

\paragraph{Evaluation of synthesis fidelity}
Image-level synthesis fidelity was evaluated using peak signal-to-noise
ratio (PSNR), which reflects voxel-wise intensity fidelity, and
three-dimensional structural similarity
($\mathrm{SSIM}_{3\mathrm{D}}$), computed between the synthetic and real
T1-weighted MRI volumes within the brain mask.

To assess preservation of anatomically meaningful structures, each synthetic
MRI and its paired real MRI were processed independently with FreeSurfer
\texttt{recon-all}. The resulting anatomical parcellations were compared
using the Dice similarity coefficient (DSC), which measures regional overlap,
and the 95th-percentile Hausdorff distance (HD95), which measures boundary
agreement. Higher PSNR, $\mathrm{SSIM}_{3\mathrm{D}}$, and DSC and lower
HD95 indicate better agreement with the real MRI.

For PSNR and $\mathrm{SSIM}_{3\mathrm{D}}$, paired subject-level comparisons
were performed using the Wilcoxon signed-rank test. One-sided tests were used
for prespecified comparisons evaluating improvement of the proposed model or
successive model components, with statistical significance defined as
$p<0.05$.

For the parcellation metrics, the proposed method was compared with each
baseline for DSC and HD95 in each of the six anatomical regions using paired
two-sided Wilcoxon signed-rank tests. Holm correction was applied across the
48 region--metric--baseline comparisons, with statistical significance defined
as an adjusted $p<0.05$.

\subsection{PET quantification}
\label{sec:methods:quant}

Real MRI-based PET quantification was used as the reference against which
the MRI-free PET quantification pipelines were evaluated. Each real or
synthetic T1-weighted MRI was processed independently with FreeSurfer
\texttt{recon-all} \citep{fischl2012freesurfer} to obtain subject-specific
anatomical parcellations. Amyloid PET uptake was then quantified using these
parcellations, with the cerebellar cortex used as the reference region
\citep{su2013quantitative}. When a synthetic MRI was used, the downstream
FreeSurfer and PET quantification procedures were kept identical to those of
the real MRI-based pipeline, with only the anatomical image replaced.

For OASIS-3, $^{11}$C-PiB PET frames acquired 40--60~min after injection were averaged to obtain a static PET volume. For each real- or synthetic-MRI-based pipeline, the PET volume was rigidly registered to the corresponding T1-weighted image, and the resulting anatomical parcellation was mapped to the native PET space using the inverse transformation and nearest-neighbor interpolation. Regional SUVRs were calculated by normalizing the mean tracer uptake in each target region to the volume-weighted mean uptake in the bilateral cerebellar cortex \citep{su2013quantitative}.

Following \citet{su2013quantitative}, the mean cortical SUVR (MCSUVR) was
calculated as the arithmetic mean of SUVRs in the precuneus, prefrontal
cortex, gyrus rectus, and lateral temporal regions. Following
\citet{kim2025automated}, regional agreement was additionally evaluated in
six cortical regions: frontal, parietal, lateral temporal, precuneus,
anterior cingulate, and posterior cingulate.

In addition to the CT-to-MRI synthesis methods, we evaluated two families of non-synthesis MRI-free PET quantification baselines, with two implementations per family: PET-only template-based normalization and CT-based anatomical normalization. The PET-only implementations used population-template-based spatial normalization \citep{edison2013,landau2023quantification}. Each PET image was nonlinearly normalized to MNI152 space using either a PiB PET template constructed from the publicly available GAAIN Centiloid calibration datasets \citep{klunk2015centiloid} or a PiB PET template constructed from OASIS-3 subjects outside the evaluation sets. These implementations are referred to as PET-template (GAAIN-derived) and PET-template (OASIS-derived), respectively. The CT-based implementations were adapted from \citet{presotto2018ct} and \citet{matsuda2021amyloid} and differed in CT preprocessing and segmentation settings. In both implementations, PET was rigidly co-registered to the same-day CT, the CT was normalized to MNI152 space using SPM unified segmentation \citep{ashburner2005unified}, and the resulting deformation was applied to PET. Target regions were defined using Desikan--Killiany-based atlases in MNI152 space \citep{desikan2006automated}. All four implementations used the same static PET image as the real MRI-based pipeline. Template and atlas construction, reference-region definitions,
and implementation-specific settings are described in
Section~\ref{sec:supp:nonsynthesis}.

Agreement with the real MRI-based reference was assessed using mean absolute
error (MAE), Pearson correlation, and Bland--Altman analysis for systematic
bias and limits of agreement. Paired subject-level absolute SUVR errors were
compared between the proposed method and the CT-to-MRI baselines using
two-sided Wilcoxon signed-rank tests, with Holm correction for multiple
comparisons. Statistical significance was defined as an adjusted
$p<0.05$.

Amyloid-positivity decision concordance was evaluated at a prespecified
MCSUVR cut-off of 1.42, selected with reference to the PiB PET threshold
reported by \citet{jack2017defining}. Decision concordance was defined
as the proportion of subjects for whom the MRI-free and real MRI-based
pipelines yielded the same amyloid-positivity classification, with
false-positive and false-negative cases reported relative to the real
MRI-based decision.

\subsection{Reliability assessment with conformal prediction}
\label{sec:methods:conformal}

\paragraph{Conformal prediction}
Even if a CT-to-MRI-based MRI-free PET quantification pipeline achieves
good overall agreement with quantification based on the real MRI, this does
not ensure that the quantification error is small for every individual
subject. In a practical MRI-free setting, the real MRI is unavailable, so
the discrepancy between an MRI-free SUVR estimate and the corresponding
real MRI-based SUVR cannot be known for a given subject. We therefore aimed
to estimate, for each subject, a range of plausible SUVR values within which
the corresponding real MRI-based SUVR would be expected to fall at a
prespecified coverage level.

Because brain CT only partially captures the anatomical information needed
for PET quantification, the difficulty of inferring the missing anatomical
information can vary across subjects. For more difficult cases,
independently trained models may converge to different plausible solutions,
leading to greater variation in the resulting SUVR estimates. We therefore
used model-to-model variation as a subject-specific uncertainty scale, with
larger variation indicating a less stable MRI-free SUVR estimate.

To quantify this uncertainty, we employed a deep ensemble
\citep{lakshminarayanan2017simple} consisting of five independently trained
instances of the proposed CT-to-MRI model, initialized differently and
trained with different mini-batch orders. Each model generated a synthetic
MRI that was independently processed through the same FreeSurfer-based PET
quantification pipeline, yielding five SUVR estimates for each subject. The
ensemble mean,
\begin{equation}
\hat{\mu}(x)
=
\frac{1}{M}\sum_{m=1}^{M} f_m(x),
\qquad M=5,
\end{equation}
was used as the SUVR estimate for the conformal analysis, and the standard deviation across
ensemble members, $\hat{\sigma}(x)$, was used as the subject-specific
uncertainty scale. (For the main synthesis and PET
quantification comparisons, the proposed method was evaluated using a single
model rather than the ensemble mean to avoid any performance gain from
ensemble averaging.)

We then calibrated this uncertainty scale using conformal prediction
\citep{vovk2005algorithmic,lei2018distribution}. For each subject $i$ in
the OASIS-3 calibration set, the normalized nonconformity score was defined
as
\begin{equation}
E_i
=
\frac{
\left|\hat{\mu}(x_i)-y_i\right|
}{
\hat{\sigma}(x_i)
},
\end{equation}
where $y_i$ denotes the corresponding real MRI-based SUVR. The score
therefore measures how large the observed MRI-free quantification error is
relative to the subject-specific ensemble variability. The distribution of
these scores in the calibration set was used to determine how much the
uncertainty scale should be expanded to achieve a desired coverage level.

For a target miscoverage level $\alpha$, the conformal scaling factor was
defined as the finite-sample-corrected empirical quantile of the calibration
scores:
\begin{equation}
\hat{q}_{1-\alpha}
=
\operatorname{Quantile}_{
\lceil(n_{\mathrm{cal}}+1)(1-\alpha)\rceil/n_{\mathrm{cal}}
}
\left(\{E_i\}_{i=1}^{n_{\mathrm{cal}}}\right),
\end{equation}
with $n_{\mathrm{cal}}=93$. For a test subject, the resulting prediction
interval for the corresponding real MRI-based SUVR was defined as
\begin{equation}
C_{\alpha}(x)
=
\left[
\hat{\mu}(x)
-
\hat{q}_{1-\alpha}\hat{\sigma}(x),
\;
\hat{\mu}(x)
+
\hat{q}_{1-\alpha}\hat{\sigma}(x)
\right].
\end{equation}

Under exchangeability between the calibration and test subjects, this
procedure provides marginal coverage of at least $1-\alpha$ for the
corresponding real MRI-based SUVR. Conformal calibration was performed
separately for MCSUVR and for each regional SUVR measure, so the coverage
guarantee applies to each measure individually rather than jointly across
all measures.

Empirical coverage was defined as the proportion of test subjects whose
real MRI-based SUVR fell within the corresponding prediction interval. We
evaluated whether the empirical coverage was consistent with the
prespecified target coverage and also measured the resulting interval width
in SUVR units. Two-sided 95\% Wilson confidence intervals were calculated for empirical coverage on the fixed test sets.

\paragraph{Reliability assessment of amyloid-positivity decisions}
We further used the conformal prediction interval to assess the reliability
of amyloid-positivity decisions based on CT-to-MRI-derived MCSUVR estimates.
Because the prediction interval represents a plausible range for the
corresponding real MRI-based MCSUVR, its position relative to the
prespecified decision cut-off $\tau=1.42$ indicates whether the decision
remains unambiguous despite uncertainty in the MRI-free estimate.

If the entire interval lay below the cut-off, the interval supported a
\textbf{confidently negative} decision. If the entire interval lay above
the cut-off, the interval supported a \textbf{confidently positive}
decision. When the interval spanned the cut-off, the corresponding real
MRI-based MCSUVR could lie on either side of the decision boundary, and the
case was therefore classified as \textbf{indeterminate}. Under this rule, a confident decision can disagree with the real MRI-based
classification only if the reference MCSUVR falls outside the prediction
interval.

We additionally examined the robustness of this decision rule to the choice
of MCSUVR cut-off by repeating the analysis across a range of cut-off values
while keeping the conformal intervals fixed. At each cut-off, we assessed
whether disagreements between the MRI-free and real MRI-based decisions were
identified as indeterminate by the conformal decision rule.

\subsection{Zero-shot external evaluation}
\label{sec:methods:external}

To evaluate zero-shot generalization, the trained CT-to-MRI models were
applied to the independent GARD cohort \citep{choi2026gwangju}, which was not used for model
training or model selection, without any retraining, fine-tuning, or
adaptation. The same preprocessing, CT-to-MRI synthesis, FreeSurfer
parcellation, and PET quantification procedures used for the internal
evaluation were applied to GARD.

CT-to-MRI synthesis and the resulting anatomical parcellations were
evaluated against the paired real T1-weighted MRI. We then evaluated
MRI-free PET quantification by comparing synthetic-MRI-based SUVR estimates
with the corresponding real MRI-based SUVR values.

We further evaluated the transferability of the conformal calibration to
the external cohort. The OASIS-3 conformal calibration was first applied directly to GARD
without recalibration, and the resulting empirical coverage was compared
with the prespecified target coverage. We additionally examined whether target-cohort recalibration could recover
the intended coverage without retraining or fine-tuning the CT-to-MRI
synthesis model. For this analysis, the synthesis models were kept fixed,
and only the conformal scaling was recalibrated using subjects from GARD.
GARD subjects were repeatedly divided into calibration and test subsets,
with the conformal scaling factor estimated from the calibration subset and
coverage and interval width evaluated in the held-out test subset.

\begin{figure}[!t]
    \centering
    \includegraphics[width=\columnwidth]{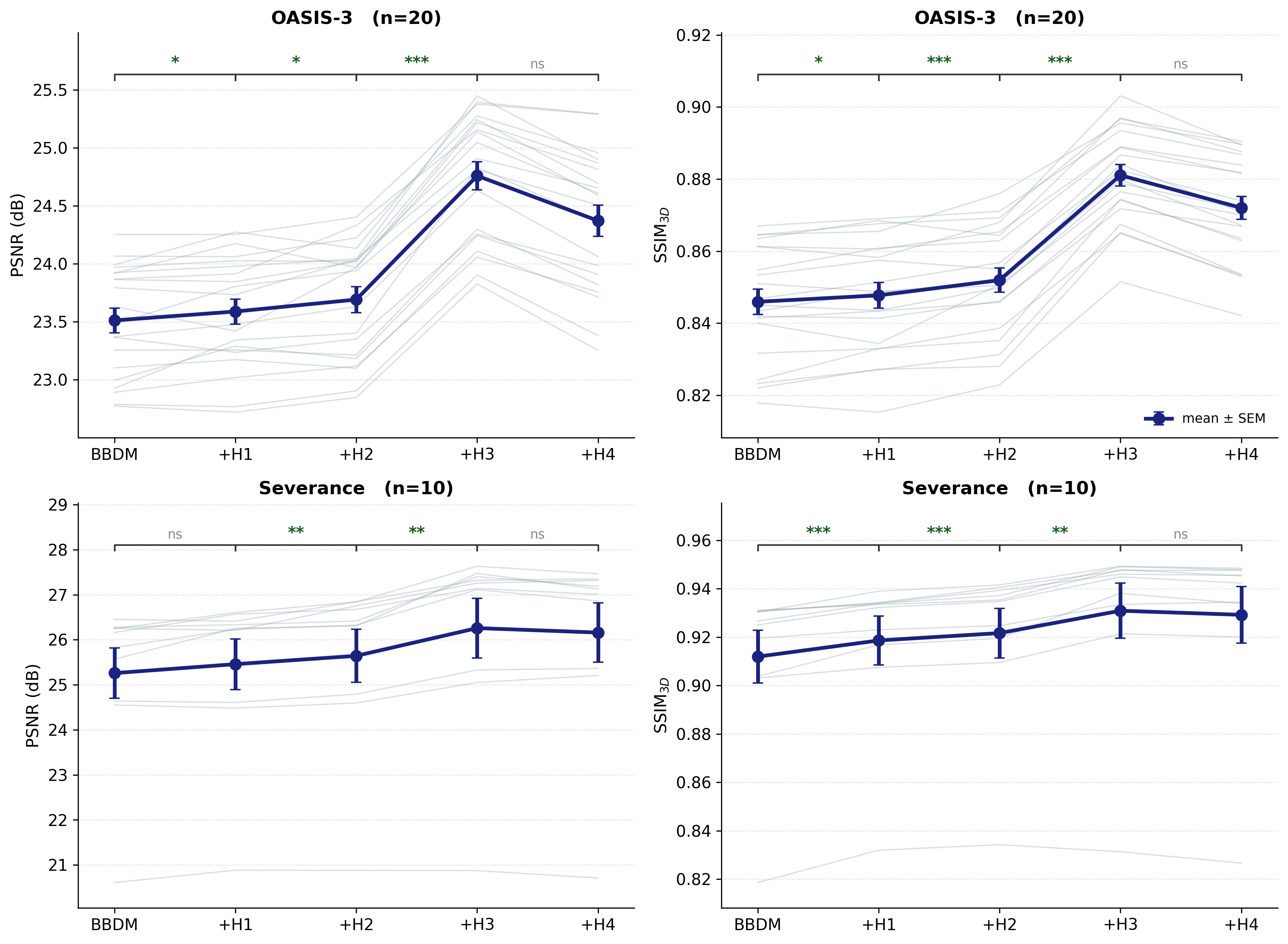}
    \caption{Cumulative ablation of the proposed design choices on the validation sets. Thin lines show paired per-subject performance, and the dark line with error bars denotes the mean $\pm$ SEM. H1 denotes multi-dataset training, H2 data-source conditioning, H3 expanded multi-slice anatomical context, and H4 segmentation-teacher feedback. Statistical significance was assessed between consecutive models using one-sided paired Wilcoxon signed-rank tests ($^{*}p<0.05$, $^{**}p<0.01$, $^{***}p<0.001$, $^\mathrm{ns}$ not significant). Because H1--H3 progressively improved synthesis performance whereas H4 provided no significant additional benefit, the final model includes H1--H3 only.}
    \label{fig:ablation}
\end{figure}

\begin{table*}[t]
\centering
\caption{Quantitative evaluation of CT-to-MRI synthesis. PSNR (dB, $\uparrow$) and $\mathrm{SSIM}_{3\mathrm{D}}$ ($\uparrow$), mean $\pm$ SD over subjects. Significance by a one-sided Wilcoxon signed-rank test against Choo2024. Best in \textbf{bold}, second-best \underline{underlined}. Our method achieves the highest fidelity on both datasets and metrics.}
\label{tab:synth}
\small
\begin{tabular}{lcccc}
\toprule
& \multicolumn{2}{c}{\textbf{Severance} ($N=22$)} & \multicolumn{2}{c}{\textbf{OASIS-3} ($N=117$)}\\
\cmidrule(lr){2-3}\cmidrule(lr){4-5}
\textbf{Method} & PSNR & $\mathrm{SSIM}_{3\mathrm{D}}$ & PSNR & $\mathrm{SSIM}_{3\mathrm{D}}$\\
\midrule
Kim2025  & $23.42 \pm 0.83$ & $0.874 \pm 0.013$ & $23.24 \pm 0.68$ & $0.807 \pm 0.018$\\
LSGAN    & $22.62 \pm 1.19$ & $0.856 \pm 0.021$ & \underline{$23.60 \pm 0.99$} & $0.840 \pm 0.020$\\
ALDM     & $21.16 \pm 1.06$ & $0.830 \pm 0.021$ & $20.95 \pm 0.89$ & $0.771 \pm 0.021$\\
Choo2024 & \underline{$25.14 \pm 1.86$} & \underline{$0.920 \pm 0.019$} & $23.48 \pm 0.82$ & \underline{$0.843 \pm 0.019$}\\
\textbf{Ours} & $\mathbf{26.02 \pm 2.04}$ & $\mathbf{0.937 \pm 0.018}$ & $\mathbf{24.69 \pm 0.96}$ & $\mathbf{0.877 \pm 0.018}$\\
\addlinespace[1pt]
\emph{$p$ vs.\ Choo2024} & $<0.001$ & $<0.001$ & $<0.001$ & $<0.001$\\
\bottomrule
\end{tabular}
\end{table*}

\begin{figure*}[t]
\centering
\includegraphics[width=0.92\linewidth]{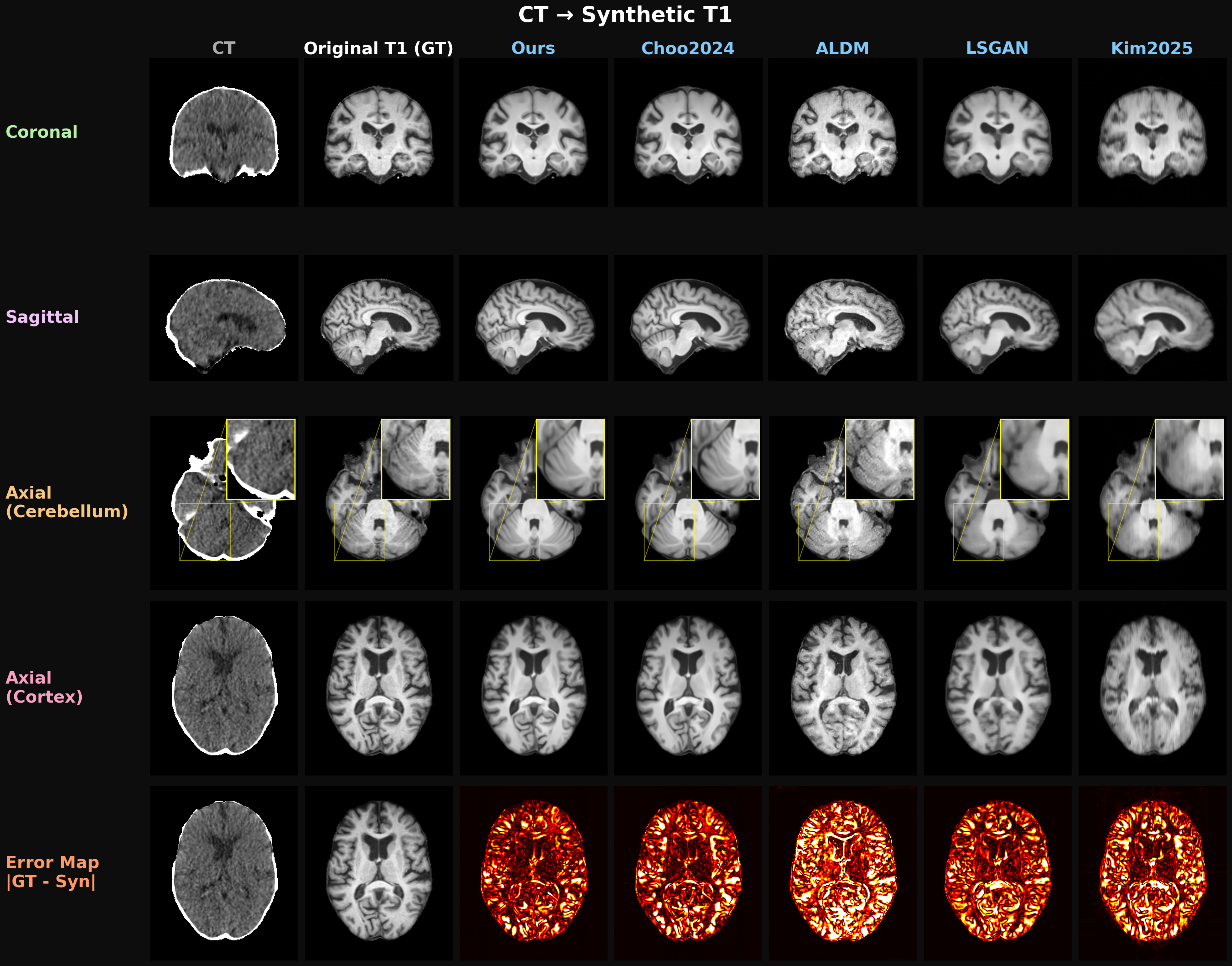}
\caption{Qualitative comparison of CT-to-MRI synthesis methods on a representative OASIS-3 subject. The columns show the CT input, ground-truth T1-weighted MRI, proposed method, and four comparison methods. The bottom row shows voxel-wise absolute error on the corresponding axial cortical slice. The proposed method shows close anatomical agreement with the ground truth, with residual errors largely localized around cortical boundaries.}
\label{fig:qual}
\end{figure*}

\section{Results}

\subsection{Compensating for limited anatomical information in CT}
\label{sec:design_results}

We first examined which design strategies could compensate for the limited anatomical information available in CT. The cumulative ablation analysis shows progressive improvement over the BBDM baseline as multi-dataset training (H1), data-source conditioning (H2), and expanded multi-slice anatomical context (H3) are introduced (Fig.~\ref{fig:ablation}; Section~\ref{sec:supp:ablation-sec}). Among these design choices, H3 provides the largest marginal performance gain, while H1 and H2 provide smaller incremental improvements. In contrast, segmentation-teacher feedback (H4) provides no significant additional improvement.

We further evaluated the extent of the multi-slice CT context by varying the number of input CT slices from 3 to 15 on the validation sets. Using a wider context than the three-slice baseline generally improves synthesis fidelity, with particularly consistent gains on OASIS-3 and performance approaching a plateau on Severance. We therefore use 15 input CT slices in the final model. Detailed results across all evaluated context sizes are provided in \ref{sec:supp:context}. Based on these analyses, the final proposed model incorporates multi-dataset training, data-source conditioning, and expanded multi-slice anatomical context.

\subsection{Anatomical fidelity of CT-to-MRI synthesis}
\begin{table*}[t]
\centering
\caption{Anatomical agreement of FreeSurfer parcellations derived from synthesized MRI with those derived from the ground-truth MRI. Results are reported as mean $\pm$ SD over OASIS-3 test subjects (N = 117 for Ours and Choo2024; N = 116 for Kim2025, LSGAN, and ALDM, each excluding one subject whose FreeSurfer parcellation failed). Dice similarity coefficient (DSC, $\uparrow$) measures regional overlap, and HD95 (mm, $\downarrow$) measures boundary agreement. Best results for each region and metric are shown in \textbf{bold}, and second-best results are \underline{underlined}.}
\label{tab:seg}
\small
\setlength{\tabcolsep}{4pt}
\begin{tabular}{llccccc}
\toprule
\textbf{Region} & \textbf{Metric} & \textbf{Kim2025} & \textbf{LSGAN} & \textbf{ALDM} & \textbf{Choo2024} & \textbf{Ours}\\
\midrule

\multirow{2}{*}{Cerebral cortex}
& DSC $\uparrow$
& $0.566 \pm 0.020$
& $0.620 \pm 0.027$
& $0.588 \pm 0.020$
& \underline{$0.655 \pm 0.025$}
& $\mathbf{0.706 \pm 0.024}$\\
& HD95 (mm) $\downarrow$
& $3.972 \pm 0.693$
& $2.379 \pm 0.382$
& \underline{$2.141 \pm 0.123$}
& $2.317 \pm 0.211$
& $\mathbf{1.677 \pm 0.228}$\\

\addlinespace[2pt]
\multirow{2}{*}{Cerebral WM}
& DSC $\uparrow$
& $0.767 \pm 0.013$
& $0.809 \pm 0.015$
& $0.774 \pm 0.015$
& \underline{$0.817 \pm 0.016$}
& $\mathbf{0.849 \pm 0.015}$\\
& HD95 (mm) $\downarrow$
& $3.755 \pm 0.358$
& $3.362 \pm 0.433$
& \underline{$2.761 \pm 0.212$}
& $2.804 \pm 0.244$
& $\mathbf{2.125 \pm 0.153}$\\

\addlinespace[2pt]
\multirow{2}{*}{Cerebellar GM (ref.)}
& DSC $\uparrow$
& $0.867 \pm 0.020$
& $0.878 \pm 0.018$
& $0.873 \pm 0.023$
& \underline{$0.891 \pm 0.019$}
& $\mathbf{0.904 \pm 0.014}$\\
& HD95 (mm) $\downarrow$
& $2.755 \pm 0.478$
& $2.538 \pm 0.371$
& $2.481 \pm 0.695$
& \underline{$2.083 \pm 0.332$}
& $\mathbf{1.873 \pm 0.302}$\\

\addlinespace[2pt]
\multirow{2}{*}{Cerebellar WM}
& DSC $\uparrow$
& $0.793 \pm 0.028$
& $0.818 \pm 0.028$
& $0.807 \pm 0.029$
& \underline{$0.825 \pm 0.034$}
& $\mathbf{0.847 \pm 0.027}$\\
& HD95 (mm) $\downarrow$
& $3.154 \pm 0.663$
& $2.975 \pm 0.528$
& $2.784 \pm 0.731$
& \underline{$2.364 \pm 0.574$}
& $\mathbf{2.047 \pm 0.475}$\\

\addlinespace[2pt]
\multirow{2}{*}{Subcortical GM}
& DSC $\uparrow$
& $0.848 \pm 0.021$
& $0.854 \pm 0.020$
& $0.858 \pm 0.025$
& \underline{$0.868 \pm 0.019$}
& $\mathbf{0.886 \pm 0.017}$\\
& HD95 (mm) $\downarrow$
& $2.032 \pm 0.514$
& $2.009 \pm 0.337$
& $1.794 \pm 0.359$
& \underline{$1.787 \pm 0.342$}
& $\mathbf{1.517 \pm 0.292}$\\

\addlinespace[2pt]
\multirow{2}{*}{Medial temporal}
& DSC $\uparrow$
& $0.796 \pm 0.032$
& $0.806 \pm 0.034$
& $0.797 \pm 0.033$
& \underline{$0.827 \pm 0.031$}
& $\mathbf{0.846 \pm 0.026}$\\
& HD95 (mm) $\downarrow$
& $2.247 \pm 0.358$
& $2.268 \pm 0.498$
& $2.211 \pm 0.384$
& \underline{$1.914 \pm 0.410$}
& $\mathbf{1.714 \pm 0.457}$\\

\bottomrule
\end{tabular}
\end{table*}

\begin{figure*}[t]
\centering
\includegraphics[width=\linewidth]{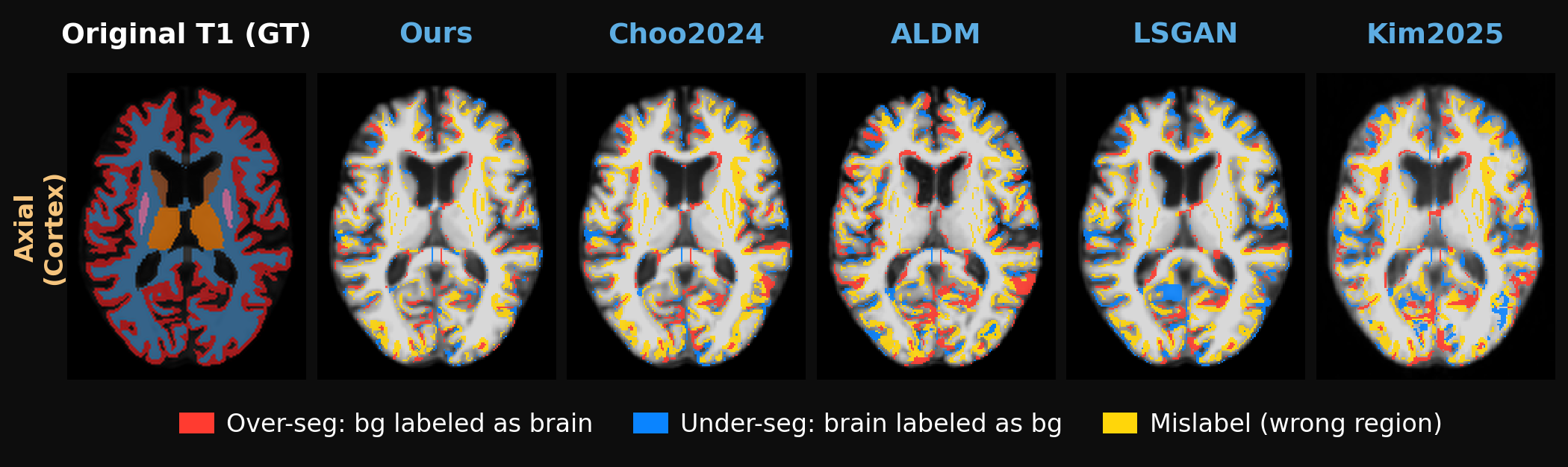}
\caption{Voxel-wise parcellation disagreement relative to the ground-truth MRI parcellation for the representative OASIS-3 subject shown in Fig.~\ref{fig:qual}. The leftmost panel shows the reference \texttt{aparc+aseg} parcellation derived from the ground-truth T1-weighted MRI. The remaining panels show disagreement for each CT-to-MRI synthesis method. Red indicates over-segmentation, blue under-segmentation, and yellow assignment to an incorrect anatomical region.}
\label{fig:segdis}
\end{figure*}

\label{sec:synth}

Having identified the final model configuration, we next evaluated its ability to reconstruct subject-specific anatomy on the Severance and OASIS-3 test sets. The proposed model achieves the highest synthesis fidelity among the competing methods (Table~\ref{tab:synth}). On the Severance cohort it attains $26.02 \pm 2.04$~dB PSNR and $0.937 \pm 0.018$ $\mathrm{SSIM}_{3\mathrm{D}}$, and on OASIS-3 $24.69 \pm 0.96$~dB and $0.877 \pm 0.018$, the best on both cohorts and both metrics. Choo2024 is the second-best method, and a one-sided Wilcoxon signed-rank test confirms that our model outperforms it with high significance ($p < 0.001$) for both metrics on both cohorts. The remaining baselines show substantially lower performance.

Qualitatively (Fig.~\ref{fig:qual}), our method and Choo2024 recover the overall cortical anatomy more faithfully than the remaining baselines, including the cerebellar folding that defines the SUVR reference region despite its weak contrast in CT. Kim2025 and LSGAN produce overly smooth cortical boundaries with substantial loss of anatomical detail. ALDM generates visually detailed MRIs, but the images contain noticeable high-frequency noise and frequently deviate from the ground truth in cortical folding patterns and other subject-specific anatomical structures. Consequently, the apparent realism of its texture does not necessarily correspond to faithful reconstruction of the individual's anatomy. Although the synthesized MRIs produced by our method and Choo2024 appear similarly realistic, the cortical error maps reveal consistently smaller residual errors for the proposed method. The remaining errors are largely confined to boundary regions around complex cortical folds.

A subject-level paired comparison against the strongest baseline, Choo2024, shows that the improvement is consistent across individual subjects. Our model outperforms Choo2024 for all subjects in $\mathrm{SSIM}_{3\mathrm{D}}$ in both cohorts and in PSNR on OASIS-3. For Severance PSNR, it outperforms Choo2024 in 21 of 22 subjects. All four one-sided Wilcoxon signed-rank tests are significant ($p<0.001$; Fig.~\ref{fig:supp:slope}).

Image-level fidelity alone does not establish whether a synthesized MRI preserves the anatomical information required for PET quantification. We therefore compared FreeSurfer parcellations derived from each synthesized MRI with those obtained from the ground-truth MRI. In the voxel-wise disagreement maps (Fig.~\ref{fig:segdis}), the proposed method shows disagreement that is largely localized around cortical boundaries and sulci. In contrast, ALDM, LSGAN, and Kim2025 exhibit more extensive disagreement across the cortical ribbon and adjacent white matter. Choo2024 shows a qualitatively similar pattern to the proposed method, but with broader disagreement in several cortical regions. Quantitatively, the proposed method achieves the highest Dice similarity coefficient (DSC) and the lowest HD95 across all six evaluated anatomical regions (Table~\ref{tab:seg}), indicating improved regional overlap and boundary agreement. Paired subject-level comparisons show
significantly better DSC and HD95 for the proposed method than for each of
the four baselines in every region, with all 48 comparisons remaining
significant after Holm correction (adjusted $p<10^{-9}$).

The improvement is particularly relevant in regions directly involved in PET quantification. In the cerebral cortex, which shows the lowest DSC among the evaluated regions, the proposed method achieves a DSC of $0.706$ and an HD95 of $1.677$~mm, compared with $0.655$ and $2.317$~mm for Choo2024, respectively. For cerebellar gray matter, which serves as the SUVR reference region, the proposed method achieves the highest DSC of $0.904$ and the lowest HD95 of $1.873$~mm. Together, these results show that the improved image-level fidelity is retained in the anatomical parcellations used for downstream PET quantification.

\subsection{MRI-free PET quantification}
\begin{table*}[tp]
\centering
\caption{Agreement between MRI-free MCSUVR estimates and the real MRI-based reference across pipelines ($N=87$). MAE is reported in SUVR units as mean $\pm$ SD, and $r$ denotes the Pearson correlation coefficient. Mean bias and 95\% limits of agreement (LoA) are derived from Bland--Altman analysis. Subject-level differences (MRI-free minus real MRI-based MCSUVR) were tested against zero using a two-sided one-sample Wilcoxon signed-rank test with Bonferroni correction, and the resulting significance is marked on the mean bias
($^{**}$adjusted $p<0.01$, $^{***}$adjusted $p<0.001$). Amyloid-positivity concordance is evaluated at the MCSUVR cut-off of 1.42, with FN and FP denoting false-negative and false-positive classifications. Best values are shown in \textbf{bold} and second-best values are \underline{underlined}. The proposed method shows the closest overall agreement with the MRI-based reference.}
\label{tab:quant}
\footnotesize
\setlength{\tabcolsep}{4pt}

\begin{tabular}{lcccccc}
\toprule
\textbf{Pipeline}
& \textbf{MAE} $\downarrow$
& \textbf{$r$} $\uparrow$
& \textbf{Mean bias}
& \textbf{95\% LoA}
& \textbf{Conc.} $\uparrow$
& \textbf{FN/FP} $\downarrow$\\
\midrule
\multicolumn{7}{l}{\textit{PET-only}}\\
PET-template (GAAIN-derived)
& $0.162 \pm 0.095$
& 0.952
& $0.119^{***}$
& $[-0.168, 0.406]$
& 89.7\%
& 0/9\\

PET-template (OASIS-derived)
& $0.131 \pm 0.080$
& 0.973
& $0.049^{**}$
& $[-0.238, 0.336]$
& 96.6\%
& 0/3\\

\addlinespace[2pt]
\multicolumn{7}{l}{\textit{CT-based}}\\
Presotto2018
& $0.055 \pm 0.051$
& 0.989
& $-0.032^{***}$
& $[-0.165, 0.101]$
& 97.7\%
& 1/1\\

Matsuda2021
& $0.053 \pm 0.062$
& 0.982
& $-0.006$
& $[-0.166, 0.155]$
& 98.9\%
& 0/1\\

\addlinespace[2pt]
\multicolumn{7}{l}{\textit{CT-to-MRI}}\\
Kim2025
& $0.044 \pm 0.033$
& 0.9955
& $0.038^{***}$
& $[-0.038, 0.115]$
& 98.9\%
& 0/1\\

LSGAN
& $0.034 \pm 0.025$
& 0.9974
& $0.029^{***}$
& $[-0.030, 0.088]$
& 98.9\%
& 0/1\\

ALDM
& $0.038 \pm 0.091$
& 0.9721
& $0.009^{**}$
& $[-0.184, 0.202]$
& 94.3\%
& 2/3\\

Choo2024
& \underline{$0.021 \pm 0.017$}
& \underline{0.9983}
& $-0.013^{***}$
& \underline{$[-0.060, 0.034]$}
& 98.9\%
& 0/1\\

\textbf{Ours}
& $\mathbf{0.016 \pm 0.015}$
& \textbf{0.9987}
& $\mathbf{-0.002}$
& $\mathbf{[-0.045, 0.040]}$
& \textbf{100\%}
& \textbf{0/0}\\
\bottomrule
\end{tabular}
\end{table*}

\begin{figure*}[tp]
\centering
\includegraphics[width=0.66\linewidth]{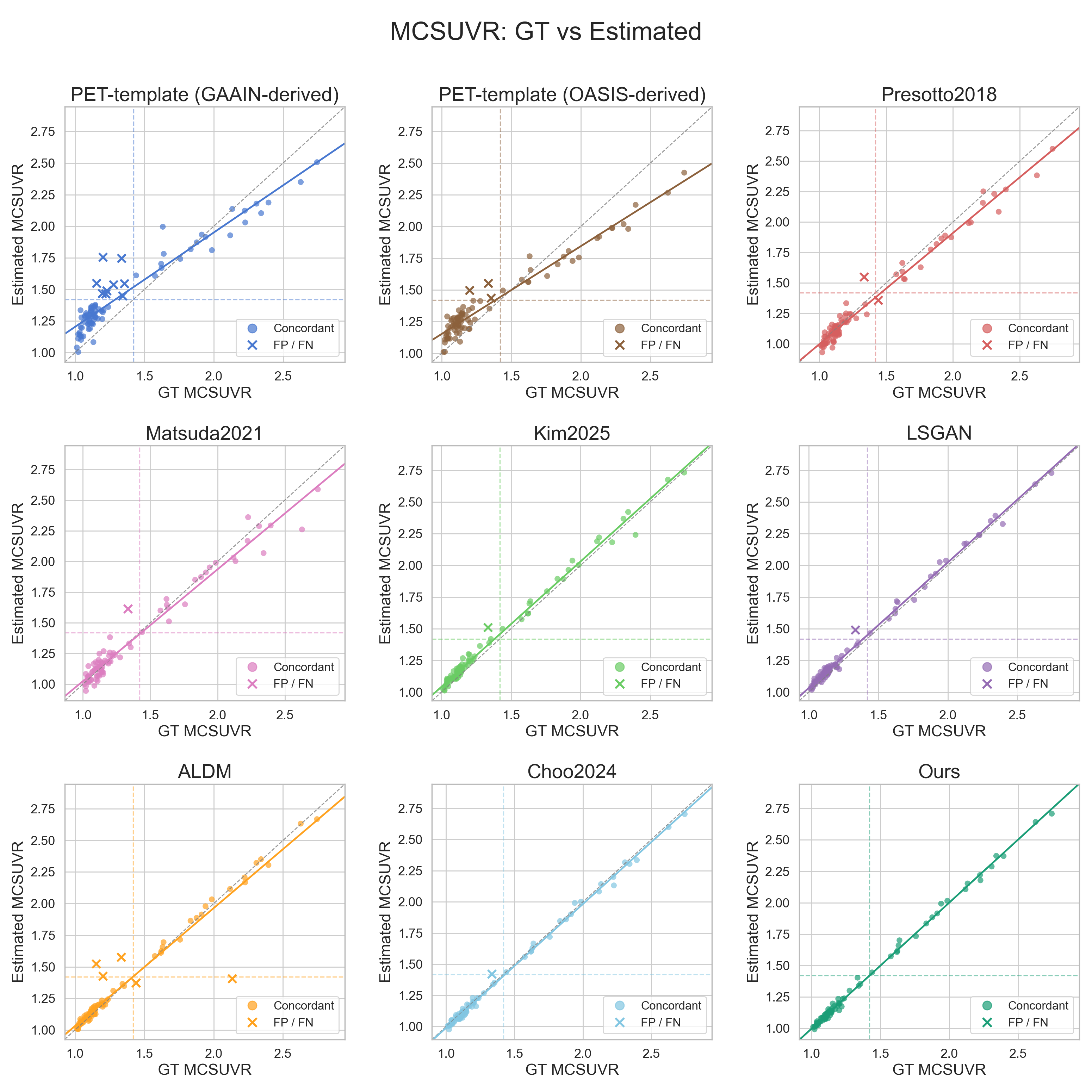}
\caption{Agreement between real MRI-based and MRI-free MCSUVR
estimates for PiB PET in the OASIS-3 test set ($N=87$).
Results are shown for two PET-only pipelines, two CT-based pipelines (Presotto2018 and Matsuda2021), and five CT-to-MRI-based pipelines.
Each point represents one subject. Crosses indicate
amyloid-positivity classification disagreements with the
MRI-based reference. The gray dashed line denotes the identity
line, the solid line represents the least-squares fit, and the
colored dashed lines mark the amyloid-positivity cut-off of 1.42. The CT-to-MRI-based pipelines, particularly the proposed method,
show the closest agreement with the real MRI-based reference,
whereas the PET-only pipelines show regression toward the cohort
mean and the CT-based pipelines show greater scatter than the
CT-to-MRI-based pipelines.}
\label{fig:mcsuvr}
\end{figure*}

\begin{table*}[t]
\centering
\caption{Mean absolute error (MAE) of regional SUVR estimates across six amyloid-relevant cortical regions. Results are reported as mean $\pm$ SD over subjects ($N=87$). Lower values indicate better agreement with the real MRI-based reference. Best values are shown in \textbf{bold}.}
\label{tab:regional_mae}
\small
\setlength{\tabcolsep}{4pt}
\begin{tabular}{lcccccc}
\toprule
\textbf{Model}
& \textbf{Frontal}
& \textbf{Parietal}
& \textbf{Lateral Temporal}
& \textbf{Precuneus}
& \textbf{Anterior Cingulate}
& \textbf{Posterior Cingulate}\\
\midrule
Kim2025
& $0.0671 \pm 0.0381$
& $0.0329 \pm 0.0340$
& $0.0352 \pm 0.0345$
& $0.0402 \pm 0.0432$
& $0.0810 \pm 0.0616$
& $0.0458 \pm 0.0454$\\

LSGAN
& $0.0432 \pm 0.0256$
& $0.0321 \pm 0.0242$
& $0.0582 \pm 0.0296$
& $0.0328 \pm 0.0344$
& $0.0466 \pm 0.0398$
& $0.0388 \pm 0.0319$\\

ALDM
& $0.0366 \pm 0.0727$
& $0.0495 \pm 0.1326$
& $0.0421 \pm 0.0801$
& $0.0574 \pm 0.1184$
& $0.0513 \pm 0.1126$
& $0.0526 \pm 0.0828$\\

Choo2024
& $0.0218 \pm 0.0154$
& $0.0249 \pm 0.0204$
& $0.0248 \pm 0.0202$
& $0.0313 \pm 0.0254$
& $0.0388 \pm 0.0309$
& $0.0284 \pm 0.0253$\\

\textbf{Ours}
& $\mathbf{0.0173 \pm 0.0148}$
& $\mathbf{0.0159 \pm 0.0148}$
& $\mathbf{0.0169 \pm 0.0172}$
& $\mathbf{0.0207 \pm 0.0229}$
& $\mathbf{0.0258 \pm 0.0200}$
& $\mathbf{0.0234 \pm 0.0215}$\\
\bottomrule
\end{tabular}
\end{table*}

\label{sec:quant}

We next examined whether the improved anatomical fidelity of the synthesized
MRI translates into more accurate PET quantification on the OASIS-3 test set
(Table~\ref{tab:quant}; Fig.~\ref{fig:mcsuvr}). Accuracy was evaluated against the real MRI-based SUVR from the same subject using the identical quantification pipeline, with only the anatomical information used for PET processing replaced. This analysis therefore measures agreement with the MRI-based reference under MRI substitution rather than diagnostic accuracy against an independent biological standard.

Among the evaluated pipelines, agreement with the MRI-based reference was lowest for the PET-only pipelines, intermediate for the CT-based pipelines, and highest for the CT-to-MRI-based pipelines. The two PET-only pipelines show the weakest agreement (GAAIN-derived template: MAE $0.162 \pm 0.095$, $r=0.952$; OASIS-derived template: MAE $0.131 \pm 0.080$, $r=0.973$). Their estimates are compressed toward the cohort mean (slopes 0.75 and 0.69), overestimating amyloid-negative subjects, with amyloid-positivity concordances of 89.7\% and 96.6\%. Incorporating subject-specific CT anatomy further reduces the error (Presotto2018: MAE $0.055 \pm 0.051$, $r=0.989$; Matsuda2021: MAE $0.053 \pm 0.062$, $r=0.982$).

The CT-to-MRI-based pipelines improve further, with the proposed method achieving the closest agreement with the MRI-based reference. It attains an MAE of ${0.016 \pm 0.015}$ and a Pearson correlation of ${r=0.9987}$, together with a mean bias of $-0.002$ and 95\% limits of agreement of $[-0.045,0.040]$. Its subject-level absolute MCSUVR errors are significantly
lower than those of all four CT-to-MRI baselines after Holm correction
(Choo2024, $p=4.5\times10^{-5}$; ALDM, $p=1.4\times10^{-4}$; LSGAN,
$p=2.4\times10^{-8}$; Kim2025, $p=4.5\times10^{-10}$). Choo2024 is the second-best method with an MAE of $0.021 \pm 0.017$ and $r=0.9983$. At the prespecified MCSUVR cut-off of 1.42, the proposed pipeline reproduces the MRI-based amyloid-positivity classification for all 87 subjects, with no false-positive or false-negative cases.

The improved agreement with the MRI-based reference is also observed across the six amyloid-relevant cortical regions (Table~\ref{tab:regional_mae}). The proposed method achieves the lowest MAE in every region, ranging from 0.0159 in the parietal cortex to 0.0258 in the anterior cingulate cortex. Regional errors are relatively larger in the precuneus and cingulate regions than in the lateral cortical regions, but remain lower than those of all competing CT-to-MRI methods. Regional errors of the non-synthesis pipelines were larger still in every
region (Supplementary Table~\ref{tab:supp:nonsynthesis_regional}).

\subsection{External generalization on the GARD cohort}
\begin{figure}[tp]
    \centering
    \includegraphics[width=\columnwidth]{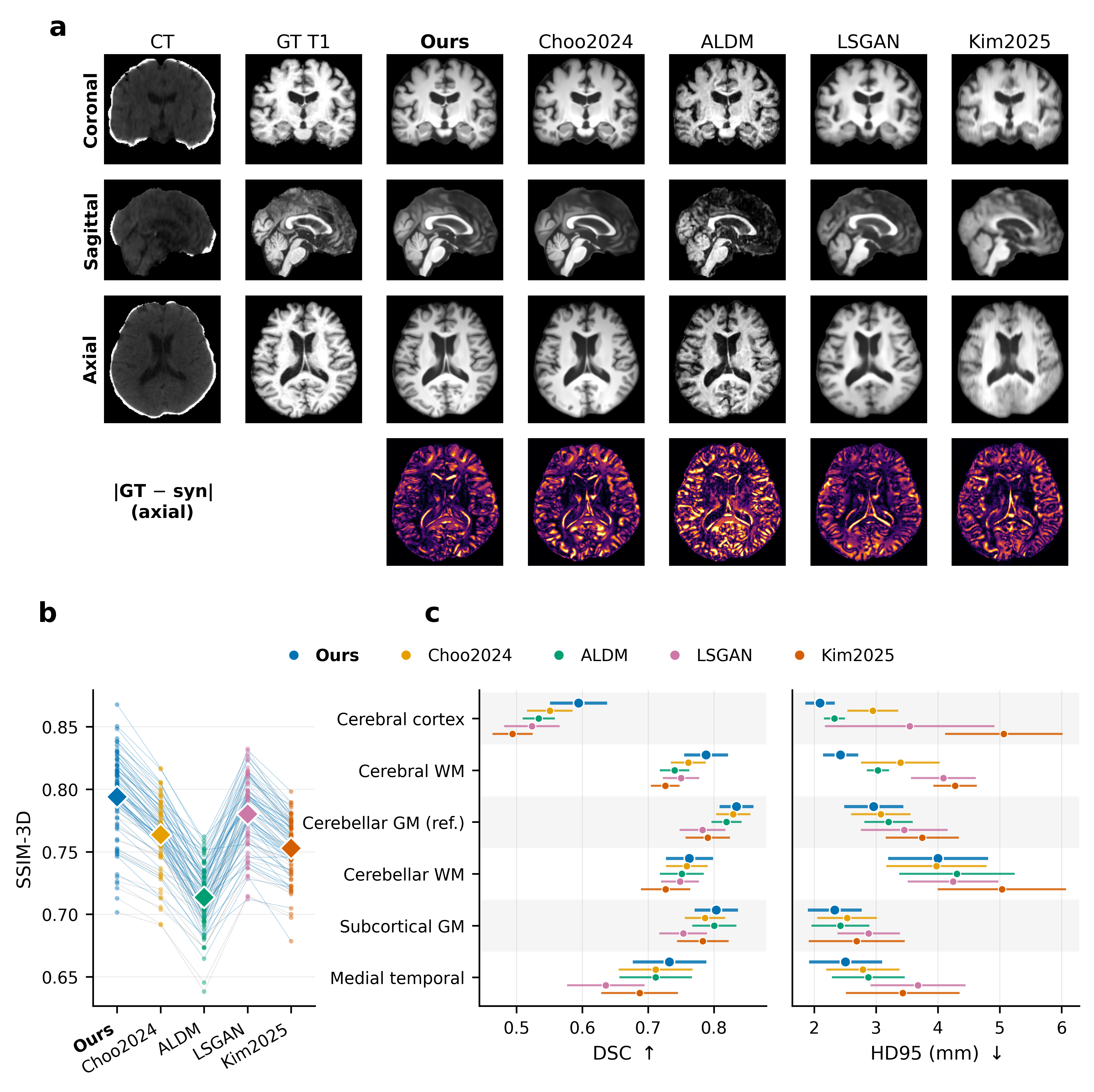}
    \caption{
    Zero-shot external evaluation of CT-to-MRI synthesis and anatomical
    parcellation on the GARD cohort.
    (a) Representative CT input, paired real T1-weighted MRI, and synthetic
    MRIs generated by the compared methods in the coronal, sagittal, and axial
    planes. The bottom row shows voxel-wise absolute error relative to the
    real MRI for the corresponding axial slice.
    (b) Subject-level $\mathrm{SSIM}_{3\mathrm{D}}$ among subjects with
    complete synthesis results ($N=84$). Thin lines connect measurements
    from the same subject across methods, and diamonds indicate the cohort
    mean. The proposed method achieves the highest
    $\mathrm{SSIM}_{3\mathrm{D}}$ in 62 of 84 subjects, while LSGAN achieves
    the highest value in the remaining 22 subjects.
    (c) Agreement between FreeSurfer parcellations derived from synthetic and
    real MRIs among subjects with valid results for all compared methods
    ($N=80$). Points and horizontal error bars indicate mean $\pm$ SD for the
    Dice similarity coefficient (DSC, $\uparrow$) and the 95th-percentile
    Hausdorff distance (HD95, mm, $\downarrow$) across six anatomical regions.
    }
    \label{fig:gard_external}
\vspace{\floatsep}
\begin{pairedfloat}{table}
\centering
\caption{
External validation of MRI-free PET quantification on the GARD cohort
($N=80$). SUVR estimates derived using each synthetic MRI are compared
with those obtained using the paired real MRI among subjects with valid
results for all compared methods. Results are reported for MCSUVR and the
same six amyloid-relevant regional SUVR measures evaluated in the primary
OASIS-3 analysis. MAE is reported in SUVR units, and $r$ denotes the
Pearson correlation coefficient. GARD was not used for model training or
adaptation. Best results for each measure are shown in \textbf{bold}, and
second-best results are \underline{underlined}.
}
\label{tab:gard_quant}

\scriptsize
\setlength{\tabcolsep}{2.4pt}
\renewcommand{\arraystretch}{1.02}

\begin{tabular}{llccccc}
\toprule
\textbf{Measure} & \textbf{Metric}
& \textbf{Ours}
& \textbf{Choo}
& \textbf{ALDM}
& \textbf{LSGAN}
& \textbf{Kim} \\
\midrule

\multirow{2}{*}{MCSUVR}
& MAE $\downarrow$ & 0.044 & 0.061 & \textbf{0.043} & \underline{0.043} & 0.052 \\
& $r$ $\uparrow$   & \textbf{0.993} & \underline{0.989} & 0.980 & 0.972 & 0.973 \\

\addlinespace[1pt]
\multirow{2}{*}{Frontal}
& MAE $\downarrow$ & \underline{0.047} & 0.068 & 0.054 & \textbf{0.042} & 0.050 \\
& $r$ $\uparrow$   & \textbf{0.993} & \underline{0.991} & 0.971 & 0.977 & 0.974 \\

\addlinespace[1pt]
\multirow{2}{*}{Parietal}
& MAE $\downarrow$ & \textbf{0.040} & 0.071 & \underline{0.053} & \underline{0.053} & 0.108 \\
& $r$ $\uparrow$   & \textbf{0.996} & \underline{0.992} & 0.985 & 0.983 & 0.975 \\

\addlinespace[1pt]
\multirow{2}{*}{Lat. temporal}
& MAE $\downarrow$ & \underline{0.033} & 0.043 & \textbf{0.032} & 0.067 & 0.048 \\
& $r$ $\uparrow$   & \textbf{0.986} & 0.973 & \underline{0.983} & 0.950 & 0.954 \\

\addlinespace[1pt]
\multirow{2}{*}{Precuneus}
& MAE $\downarrow$ & \textbf{0.053} & 0.087 & \underline{0.068} & 0.081 & 0.111 \\
& $r$ $\uparrow$   & \textbf{0.996} & \underline{0.993} & 0.987 & 0.980 & 0.980 \\

\addlinespace[1pt]
\multirow{2}{*}{Ant. cingulate}
& MAE $\downarrow$ & \textbf{0.052} & 0.073 & \underline{0.064} & 0.074 & 0.095 \\
& $r$ $\uparrow$   & \textbf{0.994} & \underline{0.987} & 0.983 & 0.983 & 0.973 \\

\addlinespace[1pt]
\multirow{2}{*}{Post. cingulate}
& MAE $\downarrow$ & \textbf{0.047} & \underline{0.066} & 0.075 & \underline{0.066} & 0.078 \\
& $r$ $\uparrow$   & \textbf{0.993} & \underline{0.985} & 0.977 & 0.982 & 0.976 \\

\bottomrule
\end{tabular}
\end{pairedfloat}
\end{figure}

\label{sec:external}

We next evaluated zero-shot external generalization on the independent
GARD cohort, which was not used for CT-to-MRI model development.
All synthesis models were applied without retraining or fine-tuning.
Despite this zero-shot setting, the proposed method maintained the
highest structural similarity to the real T1-weighted MRI among the
compared methods (Fig.~\ref{fig:gard_external}). Among the 84 subjects
with complete synthesis results, the proposed method achieved the highest
$\mathrm{SSIM}_{3\mathrm{D}}$ in 62 subjects, whereas LSGAN achieved the
highest value in the remaining 22 subjects. Overall, the proposed method
achieved an $\mathrm{SSIM}_{3\mathrm{D}}$ of $0.794 \pm 0.034$, compared
with $0.780 \pm 0.029$ for LSGAN, which was the strongest baseline.

The improved synthesis fidelity was also reflected in the anatomical
parcellations derived from the synthetic MRIs
(Fig.~\ref{fig:gard_external}). Among the 80 subjects with valid
FreeSurfer results for the real MRI and all compared methods, the proposed
method achieved the best result in 11 of the 12 region--metric
comparisons across DSC and HD95. The largest advantage was observed in
the cerebral cortex, whereas performance in cerebellar white matter was
comparable to Choo2024. Notably, although LSGAN ranked second in
$\mathrm{SSIM}_{3\mathrm{D}}$, this did not translate into comparable
FreeSurfer parcellation accuracy.

We finally examined whether the zero-shot generalization observed in
CT-to-MRI synthesis translated into MRI-free PET quantification on the
external GARD cohort (Table~\ref{tab:gard_quant}). Among the 80 subjects
with valid PET quantification results for the real MRI and all compared
methods, the proposed pipeline maintained strong agreement with the real
MRI-based reference across MCSUVR and the same six regional SUVR measures
evaluated in the primary OASIS-3 analysis. The proposed method achieved
the lowest MAE in four of the six regional SUVR measures and the highest
Pearson correlation for MCSUVR and all six regional measures. ALDM and LSGAN yielded a slightly lower MAE for MCSUVR, and ALDM also for the lateral temporal region, whereas LSGAN yielded the lowest MAE in the frontal region. Overall, the
proposed method showed the most consistent agreement across the evaluated
regional measures. Together, these results show that the proposed method achieved favorable
performance across CT-to-MRI synthesis, anatomical parcellation, and
MRI-free PET quantification when applied zero-shot to an independent
cohort.

\subsection{Reliability assessment of MRI-free PET quantification}
\begin{figure}[tp]
\begin{pairedfloat}{table}
\centering
\caption{
Empirical coverage and prediction interval width for MRI-free PET
quantification using deep-ensemble-based conformal prediction.
MCSUVR results are reported at target coverage levels of 80\%, 90\%, and 95\%,
and regional SUVR results at the 90\% target coverage.
Empirical coverage denotes the proportion of test subjects whose MRI-based
reference SUVR falls within the prediction interval, with counts shown in
parentheses as covered subjects/total subjects.
The 95\% CI column reports two-sided Wilson confidence intervals for
coverage in percentage units.
Interval width is reported in SUVR units as mean $\pm$ SD across test
subjects ($N=87$).
}
\label{tab:conformal}

\small
\setlength{\tabcolsep}{1.5pt}
\renewcommand{\arraystretch}{1.08}
\setlength{\medmuskip}{2mu}

\begin{tabular*}{\columnwidth}
{@{\extracolsep{\fill}}lcc>{\footnotesize}cc@{}}
\toprule
\textbf{Measure}
& \textbf{Target}
& \textbf{Empirical}
& \textbf{95\% CI}
& \textbf{Width} \\
\midrule

\multicolumn{5}{@{}l}{\textit{Composite MCSUVR}} \\
\addlinespace[2pt]

MCSUVR
& 80\%
& 80.5\% (70/87)
& [70.9, 87.4]
& $0.052 \pm 0.030$ \\
\addlinespace[1pt]

& 90\%
& 92.0\% (80/87)
& [84.3, 96.0]
& $0.083 \pm 0.047$ \\
\addlinespace[1pt]

& 95\%
& 96.6\% (84/87)
& [90.3, 98.8]
& $0.105 \pm 0.059$ \\

\midrule

\multicolumn{5}{@{}l}{\textit{Regional SUVR at 90\% target}} \\
\addlinespace[2pt]

Frontal
& 90\%
& 88.5\% (77/87)
& [80.1, 93.6]
& $0.094 \pm 0.049$ \\
\addlinespace[1pt]

Parietal
& 90\%
& 88.5\% (77/87)
& [80.1, 93.6]
& $0.069 \pm 0.036$ \\
\addlinespace[1pt]

\makecell[l]{Lateral\\temporal}
& 90\%
& 90.8\% (79/87)
& [82.9, 95.3]
& $0.074 \pm 0.039$ \\
\addlinespace[1pt]

Precuneus
& 90\%
& 92.0\% (80/87)
& [84.3, 96.0]
& $0.101 \pm 0.046$ \\
\addlinespace[1pt]

\makecell[l]{Anterior\\cingulate}
& 90\%
& 96.6\% (84/87)
& [90.3, 98.8]
& $0.185 \pm 0.082$ \\
\addlinespace[1pt]

\makecell[l]{Posterior\\cingulate}
& 90\%
& 96.6\% (84/87)
& [90.3, 98.8]
& $0.132 \pm 0.075$ \\

\bottomrule
\end{tabular*}
\end{pairedfloat}
\vspace{\floatsep}
    \centering
    \includegraphics[width=\columnwidth]{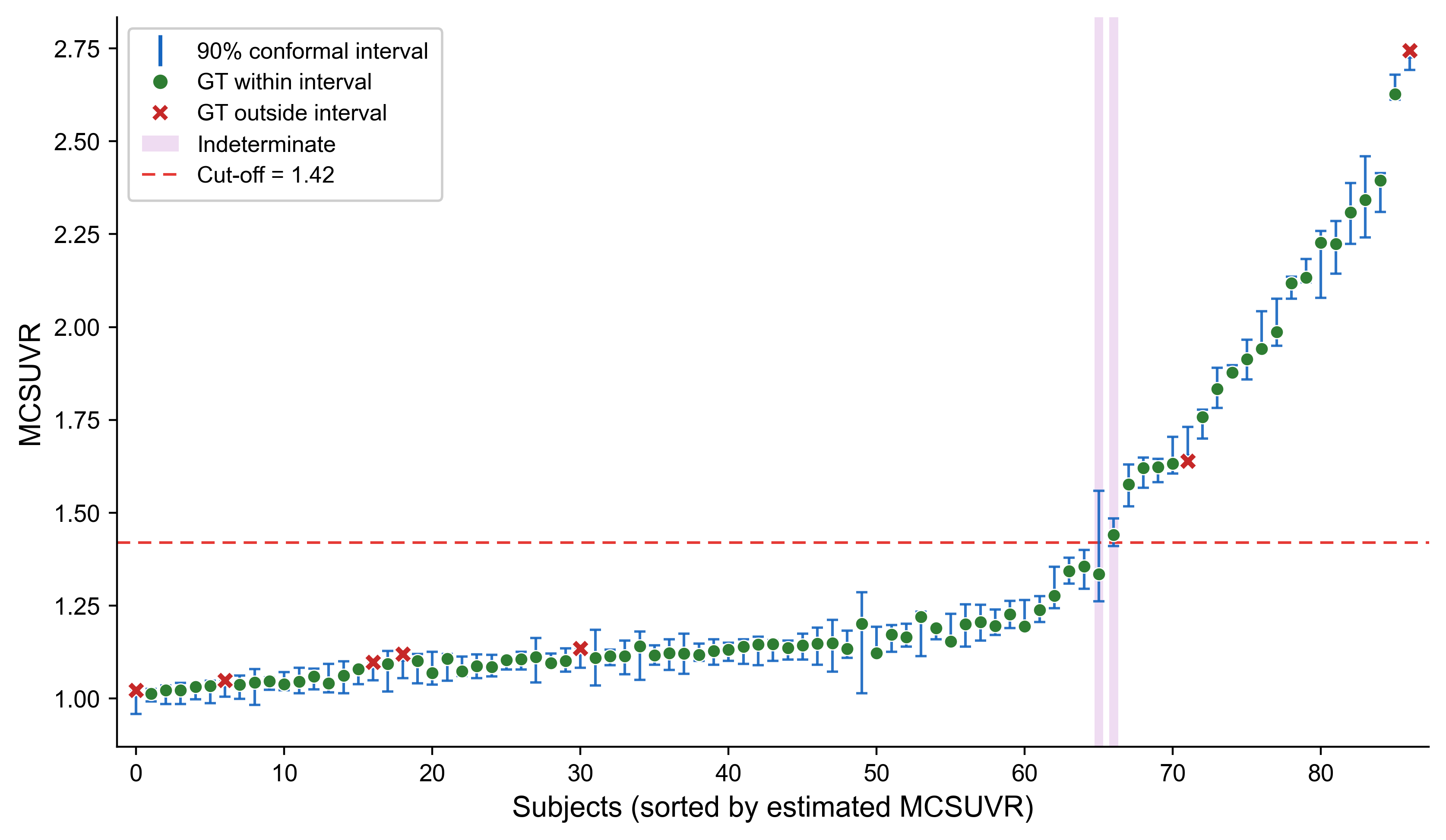}
    \caption{
    Subject-level MCSUVR conformal prediction intervals at the 90\% target coverage.
    Subjects are ordered by the ensemble-mean MRI-free MCSUVR estimate.
    Green circles and red crosses indicate whether the MRI-based reference MCSUVR falls inside or outside the interval, respectively.
    The dashed line marks the prespecified PiB PET amyloid positivity cut-off of 1.42, and shaded regions indicate indeterminate subjects whose intervals span the cut-off.
    Empirical coverage was 92.0\% (80/87), with two indeterminate subjects.
    }
    \label{fig:conformal_mcsuvr}
\end{figure}

\label{sec:reliability}

We next evaluated whether the conformal prediction intervals achieved the
prespecified coverage for the corresponding real MRI-based SUVR
(Table~\ref{tab:conformal}). For MCSUVR, empirical coverage closely matched
the target coverage across the evaluated settings. Target coverages of
80\%, 90\%, and 95\% yielded empirical coverages of 80.5\%, 92.0\%, and
96.6\%, respectively, while mean prediction interval width increased from
0.0524 to 0.1050 SUVR units with increasing target coverage.

At the 90\% target coverage, empirical coverage across the six regional SUVR measures ranged from 88.5\%
to 96.6\%. Mean prediction interval width varied across regions, from
0.0693 SUVR units in the parietal cortex to 0.1847 in the anterior
cingulate cortex (Table~\ref{tab:conformal}). The posterior cingulate also
showed relatively wide intervals, whereas the parietal and lateral temporal
regions showed the narrowest intervals. Regional empirical coverage and
prediction interval width at the 80\% and 95\% target coverage levels are
provided in \ref{sec:supp:regional_conformal}.

We next examined whether the prediction intervals could support confident
amyloid-positivity decisions in the absence of the real MRI
(Fig.~\ref{fig:conformal_mcsuvr}). For this decision analysis, we used the
90\% target coverage as the primary setting and the prespecified
MCSUVR cut-off of 1.42 as the decision boundary. 

At the 90\% target coverage, 85 of 87 subjects received a confident decision,
including 65 confident negatives and 20 confident positives, while two
subjects were designated indeterminate. These two subjects had ensemble-mean
MCSUVR estimates of 1.4103 and 1.4473, both close to the 1.42 cut-off. In both
cases, the corresponding prediction intervals crossed the decision boundary,
indicating that the unknown real MRI-based SUVR could plausibly fall on either
side. Both point estimates were nevertheless concordant with the amyloid-
positivity classification based on the real MRI.

At the prespecified MCSUVR cut-off of 1.42, the amyloid-positivity decisions
obtained from our synthetic-MRI-based MRI-free pipeline reproduced the real
MRI-based decisions for all 87 test subjects, with no false-positive or
false-negative cases. We further examined whether this agreement was
sensitive to the choice of MCSUVR cut-off and whether conformal prediction
could identify unreliable decisions when disagreements emerged. In a
supplementary sensitivity analysis, the cut-off was varied from 1.30 to 1.50
while the 90\% conformal prediction intervals were held fixed. Disagreements
between the MRI-free and real MRI-based decisions occurred only at cut-offs
from 1.34 to 1.41. All such cases were designated indeterminate because their
prediction intervals spanned the cut-off. Conversely, all decisions retained
as confident remained concordant with the real MRI-based reference across the
evaluated cut-off range (\ref{sec:supp:mcsuvr_threshold}). These results
indicate that the conformal intervals can distinguish cases requiring caution
from those for which the MRI-free amyloid-positivity decision remains stable
with respect to the real MRI-based reference.

Finally, we examined whether conformal calibration performed on OASIS-3
transferred to the independent GARD cohort. Although MRI-free SUVR
estimates retained strong zero-shot agreement with the MRI-based reference
on GARD, prediction intervals calibrated on OASIS-3 substantially
undercovered the GARD reference values. For MCSUVR, empirical coverage at
the 90\% target level decreased to 62.2\%. In contrast, when the same
conformal procedure was recalibrated within GARD using repeated random
split-half calibration/test partitions, empirical coverage returned close
to the target level ($90.7 \pm 6.2$\% across 1,000 splits). Similar patterns
were observed across the six regional SUVR measures
(\ref{sec:supp:gard_conformal}). These results indicate that conformal
calibration did not transfer directly across cohorts, but could be restored
using a calibration subset from the target cohort without retraining or
fine-tuning the CT-to-MRI synthesis model.

\section{Discussion}

In this study, we addressed three questions central to CT-to-MRI-based
MRI-free PET quantification. First, how can subject-specific T1-weighted MRI
anatomy be inferred more faithfully from the limited anatomical information
available in brain CT? Second, does improving this anatomical inference
translate into more accurate MRI-free PET quantification? Third, in a
practical setting where the real MRI is unavailable, how can the reliability
of an individual MRI-free SUVR estimate be assessed?

To address the first question, we strengthened the anatomical information
available to the synthesis model through multi-dataset training,
data-source conditioning, and expanded multi-slice CT context. The resulting
improvements in CT-to-MRI synthesis were accompanied by more accurate
anatomical parcellation and closer agreement with real MRI-based PET
quantification. To address the third question, we combined deep-ensemble
uncertainty with conformal prediction to estimate a plausible range for the
corresponding real MRI-based SUVR and used this range to assess the
reliability of amyloid-positivity decisions. Finally, we evaluated whether
these improvements generalized to an independent external cohort without
retraining or adaptation of the CT-to-MRI synthesis model.

\paragraph{Improving MRI-free PET quantification through better anatomical inference}
Previous work \citep{kim2025automated} showed that CT-to-MRI synthesis
can be incorporated into MRI-based PET quantification pipelines, establishing
its feasibility for MRI-free PET quantification. Separately,
\citet{choo2024slice} improved CT-to-MRI synthesis using a slice-consistent
Brownian Bridge diffusion framework \citep{li2023bbdm}. Against this
background, we examined whether richer anatomical information available to
the synthesis model could further improve CT-to-MRI synthesis and, in turn,
MRI-free PET quantification. Multi-dataset training
and data-source conditioning provided incremental improvements, whereas
expanded multi-slice CT context contributed the largest gain in the ablation
analysis. These findings suggest that improving CT-to-MRI synthesis depends
not only on the generative formulation, but also on the anatomical prior
learned from the training data and the amount of subject-specific anatomical
information available from the input CT.

These improvements were reflected in downstream PET quantification. Among the evaluated pipelines, agreement with the real MRI-based reference was lowest for the PET-only pipelines, intermediate for the CT-based pipelines, and highest for the CT-to-MRI-based pipelines.
The proposed method achieved the closest agreement with the MRI-based
reference on OASIS-3 across both composite and regional SUVR measures. This is
particularly relevant because accurate SUVR quantification depends on
anatomical localization of both cortical target regions and the cerebellar
reference region \citep{su2013quantitative}.

The FreeSurfer parcellation analysis further showed that the improvement was not limited
to image-level similarity. Although visually plausible synthetic images do
not necessarily preserve all subject-specific anatomical features
\citep{cohen2018distribution}, the proposed method also improved regional
overlap and boundary agreement of the derived anatomical parcellations. This
supports the interpretation that the improved PET quantification was
associated with better preservation of anatomy relevant to regional PET
sampling rather than image appearance alone.

Residual errors nevertheless varied across cortical regions. Errors were
relatively larger in the precuneus and anterior and posterior cingulate
cortices than in the lateral cortical regions. These medial regions contain
complex cortical folding and narrow gray--white matter boundaries that are
only weakly represented in CT, making their individual anatomy more difficult
to infer. Small boundary differences in these structures can also alter
regional PET quantification. The persistence of larger errors in these regions
therefore indicates that recovery of fine medial cortical anatomy remains an
important challenge for CT-based MRI inference.

Although the present study focused on amyloid PET, the overall framework is
not inherently tracer-specific. MRI-derived anatomical information is also
used for regional quantification of other brain PET tracers, including
$^{18}$F-FDG PET and tau PET
\citep{bailly2015precuneus,mishra2017av}. Accordingly, the same principle
could potentially be extended to these applications by replacing the
unavailable MRI with CT-derived synthetic MRI while retaining the
tracer-specific PET quantification pipeline. Validation across additional
tracers and clinical applications will be necessary to establish this
broader applicability.

\paragraph{Estimating the plausible range of the real MRI-based SUVR}
Accurate MRI-free PET quantification does not by itself indicate how close an
individual SUVR estimate is to the value that would have been obtained using
the real MRI. In a practical MRI-free setting, the real MRI is unavailable,
so this discrepancy cannot be directly measured for a given subject.
Therefore, even when MRI-free SUVR estimates show high overall agreement with
the real MRI-based reference, it remains important to estimate a plausible
range for the corresponding real MRI-based SUVR for each subject.

We addressed this problem by constructing calibrated prediction intervals for
the corresponding real MRI-based SUVR using deep-ensemble uncertainty and
conformal prediction
\citep{lakshminarayanan2017simple,vovk2005algorithmic,
lei2018distribution,angelopoulos2023gentle}. Across the evaluated target
coverage levels, empirical coverage closely followed the intended coverage
for MCSUVR, with greater variation across regional SUVR measures.
The coverage confidence intervals reflect the limited precision of these
estimates from 87 test subjects. These results support the feasibility of
estimating a plausible range for the corresponding real MRI-based SUVR
within OASIS-3, where the calibration and test subjects were drawn from
the same cohort, even when the real MRI itself is unavailable.

The width of this interval adds information that is not available from the
MRI-free SUVR value alone. A narrow interval constrains the possible real
MRI-based SUVR to a relatively small range around the MRI-free estimate,
whereas a wider interval indicates that a broader range of real MRI-based
SUVR values remains compatible with the MRI-free result. Such information
may help determine whether an MRI-free quantification result is sufficiently
reliable for its intended use or whether further evaluation, such as
acquisition of an MRI, should be considered.

This becomes particularly relevant when the SUVR is interpreted relative to
an amyloid-positivity decision cut-off. At the primary 90\% target coverage,
85 of 87 subjects (97.7\%) had prediction intervals that remained entirely on
one side of the prespecified cut-off, whereas only two subjects were
classified as indeterminate. In this retrospective cohort, this suggests that
the large majority of subjects could have received an unambiguous
amyloid-positivity decision from the MRI-free pipeline, while additional
assessment could have been reserved for the small subset whose prediction
intervals crossed the decision boundary. The sensitivity analysis further supported this interpretation, as all
observed disagreements between the MRI-free and real MRI-based decisions
were captured by the indeterminate category across the evaluated cut-offs.
In interpreting these results, the nominal coverage level applies to the
prediction intervals and does not directly specify the classification
accuracy among subjects receiving a confident decision.

If confirmed prospectively, such an accept-or-defer strategy could have
practical implications for the use of MRI in PET quantification. Rather than
requiring an additional MRI for every subject, MRI or another confirmatory
assessment could potentially be prioritized for cases in which the MRI-free
SUVR remains insufficiently constrained or its prediction interval overlaps a
clinically relevant decision boundary. This could reduce additional imaging
burden and potentially lower the resources required for quantitative PET
assessment, while preserving a mechanism for referring uncertain cases for
further evaluation. A formal cost-effectiveness analysis would, however, be
required before drawing conclusions regarding economic benefit.

This selective strategy may also be relevant for patients in whom MRI is
difficult or unavailable because of contraindications, claustrophobia,
motion, acquisition burden, or other practical limitations
\citep{nguyen2020prevalence}. Previous MRI-free approaches have similarly
sought to reduce dependence on structural MRI for PET quantification
\citep{edison2013,presotto2018ct,matsuda2021amyloid,kim2025automated}.
Providing both an MRI-free SUVR and an estimate of its plausible range may
extend this goal by allowing quantitative PET assessment even when MRI cannot
readily be obtained, while explicitly identifying subjects for whom the
MRI-free result remains too uncertain to rely on without further evaluation.

\paragraph{Zero-shot external generalization and conformal calibration transfer}
The independent GARD evaluation examined whether the proposed approach could
be applied beyond the datasets used for model training. Despite differences
in CT acquisition characteristics and amyloid tracer, the proposed model
largely preserved its relative ranking across CT-to-MRI synthesis, anatomical
parcellation, and MRI-free PET quantification without retraining or
fine-tuning of model parameters, although its absolute errors were larger
than in OASIS-3. These findings provide preliminary evidence that the learned
CT-to-MRI mapping may generalize across cohorts and acquisition settings.

In contrast, direct transfer of the conformal calibration from OASIS-3 to
GARD resulted in substantial undercoverage, suggesting a shift in the
distribution of ensemble-normalized prediction errors between cohorts.
Because standard conformal prediction relies on exchangeability between
calibration and test samples, such distributional differences can invalidate
the nominal coverage guarantee
\citep{tibshirani2019conformal,angelopoulos2023gentle}.
Differences in cohort characteristics, imaging acquisition, and amyloid
tracer may all have contributed to this shift.

Notably, empirical coverage returned close to the target level after
recalibrating the conformal scaling using a subset of GARD subjects, while
the CT-to-MRI synthesis models themselves were left unchanged. In a new
cohort, it may therefore not be necessary to retrain the synthesis model
solely to recover the intended prediction-interval coverage. Instead,
conformal calibration can be updated using a target-cohort calibration set
while retaining the existing synthesis model. In the GARD analysis, this
required paired CT, real T1-weighted MRI, and amyloid PET data from 41
subjects per split. At the 90\% target coverage, mean MCSUVR interval width
on GARD increased from 0.119 with OASIS-3 calibration to 0.203 with
GARD-specific recalibration.
This suggests the possibility that, after local recalibration with a limited
number of paired cases, the approach could be applied to a new imaging
environment and still provide a plausible range for the corresponding real
MRI-based SUVR in subsequent subjects for whom MRI is not available.

\paragraph{Limitations and future directions}
Several challenges remain before the proposed approach can be made more
generalizable and clinically robust. First, the anatomical information
available in CT is fundamentally limited, particularly for fine gray--white
matter boundaries and complex cortical folding. The relatively larger errors
observed in the precuneus and cingulate regions suggest that subject-specific
anatomy cannot always be recovered accurately from CT alone. Further work is
therefore needed to improve the recovery of fine individual anatomy from the
limited structural information available in CT.

Second, the current data-source conditioning strategy limits straightforward
application to previously unseen imaging environments. The model was trained
using discrete source labels corresponding to the datasets included during
training, whereas a new external dataset does not have a predefined source
label. For GARD, the OASIS-3 label was used as a surrogate because its CT
characteristics were most similar, but this is not a general solution. More
generalizable conditioning strategies that can infer acquisition or image
characteristics directly from the input CT, rather than relying on predefined
dataset identities, will therefore be needed for broader deployment.

Third, the present training data were not designed to systematically
represent focal or atypical brain abnormalities, such as brain tumors or
substantial white matter lesions. Medical image translation models can alter
or fail to preserve pathological features that are uncommon in the training
distribution \citep{cohen2018distribution}. In the present pipeline, such
errors could propagate beyond the synthesized image, because inaccurate
representation of abnormal anatomy may affect anatomical parcellation and
regional PET sampling, and consequently alter the resulting SUVR. Future work
should therefore develop synthesis models that can more reliably preserve
such structural abnormalities and prevent their errors from propagating to
anatomical parcellation and PET quantification.

Fourth, the PET analyses were restricted to subjects with same-day PiB PET and T1-weighted MRI and to scans passing quality control, which may favor higher-quality acquisitions. In addition, differences in synthesis or quantification performance across age, sex, or ethnic groups were not assessed, although the development (OASIS-3) and external (GARD) cohorts
differ in population.

Finally, this study was retrospective and external validation was performed
in a single additional cohort. Prospective multi-center validation across
broader scanners, acquisition protocols, populations, and PET tracers will
be needed to establish the robustness and generalizability of the proposed
approach.
\section{Conclusion}

In conclusion, this study improved the accuracy of CT-to-MRI-based MRI-free PET quantification and showed that, when conformal prediction is calibrated for the target cohort, the reliability of individual SUVR estimates can be assessed when the real MRI is unavailable. By strengthening the anatomical
prior learned during training and incorporating richer subject-specific
information from the input CT, the proposed approach improved CT-to-MRI
synthesis, anatomical parcellation, and downstream PET quantification relative to the other approaches evaluated here. The resulting MRI-free SUVR estimates showed close agreement with the corresponding real MRI-based reference in OASIS-3 and favorable performance
when applied zero-shot to an independent external cohort.

Beyond improving quantification accuracy, the proposed reliability framework
used deep-ensemble uncertainty and conformal prediction to estimate a
plausible range for the corresponding real MRI-based SUVR. In OASIS-3, this made it possible to distinguish cases in which an MRI-free
amyloid-positivity decision was well separated from the decision cut-off
from those in which additional evaluation may be warranted. The external evaluation further showed that
conformal calibration may not transfer directly across cohorts, but that
target-cohort recalibration can restore the intended coverage without
retraining the CT-to-MRI synthesis model. Together, these findings support the
potential of CT-to-MRI-based MRI-free PET quantification to provide both
accurate SUVR measurements and an explicit assessment of their reliability,
while highlighting the importance of robust anatomical inference and local
calibration for broader clinical deployment.

\section*{CRediT authorship contribution statement}
\textbf{Jungwoo Seo:} Conceptualization, Methodology, Software, Formal analysis, Investigation, Data curation, Writing -- original draft, Visualization.
\textbf{Seung Yun Choi:} Software, Validation, Data curation, Writing -- review \& editing.
\textbf{Kun Ho Lee:} Resources, Data curation.
\textbf{Seun Jeon:} Conceptualization, Methodology, Resources, Data curation, Writing -- review \& editing, Supervision.
\textbf{Byoung Seok Ye:} Conceptualization, Resources, Data curation, Writing -- review \& editing, Supervision, Funding acquisition.
\textbf{Jiook Cha:} Conceptualization, Methodology, Resources, Writing -- review \& editing, Supervision, Project administration, Funding acquisition.

\section*{Declaration of competing interest}
The authors declare no competing interests.
 
\section*{Data and code availability}
Code will be made publicly available upon acceptance of this
manuscript at
\url{https://github.com/Transconnectome/CT-to-MRI-PETQ-UQ}.
Key implementation settings are summarized in
Table~\ref{tab:supp:impl}.
OASIS-3 data are available through
\url{https://www.oasis-brains.org}, subject to the OASIS
Data Use Agreement.
SynthRAD2023 data are available at
\url{https://doi.org/10.5281/zenodo.7260705}
under the CC BY-NC 4.0 license.
The Severance Hospital imaging data used in this study
are not publicly available due to patient privacy
protections and institutional data-use restrictions
prohibiting third-party sharing.
The GARD imaging data used in this study are not
publicly available and cannot be redistributed by
the authors. Inquiries regarding potential access
should be directed to the GARD Cohort Research Center
(\href{mailto:GARD@chosun.ac.kr}{GARD@chosun.ac.kr}).

\section*{Ethics}
The retrospective analysis of the Severance Hospital
clinical and imaging data was approved by the
Institutional Review Board of Yonsei University Health
System, Severance Hospital (approval no.~4-2025-1362).
The requirement for informed consent was waived by
the Institutional Review Board.
The published GARD cohort description reports approval
by the Institutional Review Boards of Chosun University Hospital and Chonnam National University Hospital. It also reports that participants or their legal guardians provided written informed consent before cohort enrollment \citep{choi2026gwangju}.
OASIS-3 and SynthRAD2023 data were used in accordance
with their respective data-use agreement and license.

\section*{Acknowledgements}

This work was supported by the National Research Foundation of Korea (NRF) grants funded by the Korea government (MSIT) (Nos. 2021R1C1C1006503, RS-2023-00266787, RS-2023-00265406, RS-2024-00342301, RS-2026-25490183, RS-2026-25524667, RS-2026-25518389, RS-2021-NR061370, NRF-2021M3E5D2A01022515, NRF-2021S1A3A2A02090597, RS-2025-18362970, and RS-2026-25554718), and by the Researchers Program through Seoul National University (Nos. 200-20250071, 200-20250049, 200-20240057, 0670-20260027, 200-20260009, 200-20250116, 200-20260081, 200-20250115, 200-20250113, and 0670-20250039).

Additional support was provided by the Institute of Information \& Communications Technology Planning \& Evaluation (IITP) grant funded by the Korea government (MSIT) [No. RS-2021-II211343, Artificial Intelligence Graduate School Program, Seoul National University], by the Global Research Support Program in the Digital Field (RS-2024-00421268), and by the Artificial Intelligence Industrial Convergence Cluster Development Project funded by the Ministry of Science and ICT and Gwangju Metropolitan City.

This work was also supported by the Korea Brain Research Institute (KBRI) Basic Research Program funded by the Ministry of Science and ICT (Nos. 25-BR-05-01 and 26-BR-05-01); by the Korea Health Industry Development Institute (KHIDI) and the Ministry of Health and Welfare, Republic of Korea (HR22C1605); by a grant for the Korea Health Technology R\&D Project through KHIDI, funded by the Ministry of Health \& Welfare, Republic of Korea (RS-2022-KH129543); by a grant of the Korean ARPA-H Project through KHIDI, funded by the Ministry of Health \& Welfare, Republic of Korea (RS-2025-25457239); by the Korea Basic Science Institute (National Research Facilities and Equipment Center) grant funded by the Ministry of Education (RS-2024-00435727); and by the Technology Innovation Program (Industrial Strategic Technology Development Program) (RS-2024-00433283, Development of an AI-Driven Platform for Clinical Validation in Senior Healthcare Focused on Degenerative Brain Diseases), funded by the Ministry of Trade, Industry \& Resources (MOTIR), Republic of Korea.

We acknowledge the National Supercomputing Center for providing supercomputing resources and technical support (KSC-2023-CRE-0568, KSC-2024-CRE-0198, and KSC-2025-CRE-0340). An award for computer time was provided by the Advanced GPU Utilization Support Program funded by the Government of the Republic of Korea (Ministry of Science and ICT) and by the U.S. Department of Energy's (DOE) ASCR Leadership Computing Challenge (ALCC). This research used resources of the National Energy Research Scientific Computing Center (NERSC), a DOE Office of Science User Facility, under ALCC award m4750-2024, as well as supporting resources at the Argonne and Oak Ridge Leadership Computing Facilities, U.S. DOE Office of Science user facilities at Argonne National Laboratory and Oak Ridge National Laboratory.

\section*{Declaration of generative AI and AI-assisted technologies in the manuscript preparation process}
During the preparation of this work, the authors used Claude (Anthropic) and ChatGPT (OpenAI) to assist with drafting and revising the manuscript text, and OpenAI's gpt-image model (version~2.0) to generate the schematic conceptual illustrations in Figs.~\ref{fig:overview} and~\ref{fig:arch}. After using these tools, the authors reviewed and edited the content as needed and take full responsibility for the content of the published article. The authors verified the scientific content, numerical results, and cited references, and no image representing study data was generated or altered using these tools.

\bibliographystyle{elsarticle-num-names}
\bibliography{references}

@book{vovk2005algorithmic,
  title     = {Algorithmic Learning in a Random World},
  author    = {Vovk, Vladimir and Gammerman, Alexander and Shafer, Glenn},
  year      = {2005},
  publisher = {Springer}
}

@article{villain2012regional,
  author  = {Villain, Nicolas and Ch{\'e}telat, Ga{\"e}l and
             Grassiot, Beno{\^i}t and Bourgeat, Pierrick and
             Jones, Gareth and Ellis, Kathryn A. and Ames, David and
             Martins, Ralph N. and Eustache, Francis and
             Salvado, Olivier and Masters, Colin L. and
             Rowe, Christopher C. and Villemagne, Victor L. and
             {AIBL Research Group}},
  title   = {Regional dynamics of amyloid-$\beta$ deposition in healthy
             elderly, mild cognitive impairment and Alzheimer's disease:
             a voxelwise {PiB--PET} longitudinal study},
  journal = {Brain},
  year    = {2012},
  volume  = {135},
  number  = {7},
  pages   = {2126--2139},
  doi     = {10.1093/brain/aws125}
}

@article{jack2024revised,
  author  = {Jack, Clifford R., Jr. and Andrews, Scott J. and
             Beach, Thomas G. and Buracchio, Teresa and Dunn, Billy and
             Graf, Ana and Hansson, Oskar and Ho, Carole and
             Jagust, William and McDade, Eric and Molinuevo, Jose Luis and
             Okonkwo, Ozioma C. and Pani, Luca and Rafii, Michael S. and
             Scheltens, Philip and Siemers, Eric and Snyder, Heather M. and
             Sperling, Reisa and Teunissen, Charlotte E. and Carrillo, Maria C.},
  title   = {Revised criteria for diagnosis and staging of
             Alzheimer's disease: Alzheimer's Association Workgroup},
  journal = {Alzheimer's \& Dementia},
  year    = {2024},
  volume  = {20},
  number  = {8},
  pages   = {5143--5169},
  doi     = {10.1002/alz.13859}
}

@article{bateman2012clinical,
  title={Clinical and biomarker changes in dominantly inherited Alzheimer's disease},
  author={Bateman, Randall J and Xiong, Chengjie and Benzinger, Tammie LS and Fagan, Anne M and Goate, Alison and Fox, Nick C and Marcus, Daniel S and Cairns, Nigel J and Xie, Xianyun and Blazey, Tyler M and others},
  journal={New England Journal of Medicine},
  volume={367},
  number={9},
  pages={795--804},
  year={2012},
  publisher={Mass Medical Soc}
}

@article{mehrtash2020confidence,
  title={Confidence calibration and predictive uncertainty estimation for deep medical image segmentation},
  author={Mehrtash, Alireza and Wells, William M and Tempany, Clare M and Abolmaesumi, Purang and Kapur, Tina},
  journal={IEEE transactions on medical imaging},
  volume={39},
  number={12},
  pages={3868--3878},
  year={2020},
  publisher={IEEE}
}

@article{tanno2021uncertainty,
  title={Uncertainty modelling in deep learning for safer neuroimage enhancement: Demonstration in diffusion MRI},
  author={Tanno, Ryutaro and Worrall, Daniel E and Kaden, Enrico and Ghosh, Aurobrata and Grussu, Francesco and Bizzi, Alberto and Sotiropoulos, Stamatios N and Criminisi, Antonio and Alexander, Daniel C},
  journal={NeuroImage},
  volume={225},
  pages={117366},
  year={2021},
  publisher={Elsevier}
}

@article{olsson2022estimating,
  title={Estimating diagnostic uncertainty in artificial intelligence assisted pathology using conformal prediction},
  author={Olsson, Henrik and Kartasalo, Kimmo and Mulliqi, Nita and Capuccini, Marco and Ruusuvuori, Pekka and Samaratunga, Hemamali and Delahunt, Brett and Lindskog, Cecilia and Janssen, Emiel AM and Blilie, Anders and others},
  journal={Nature communications},
  volume={13},
  number={1},
  pages={7761},
  year={2022},
  publisher={Nature Publishing Group UK London}
}

@article{choo2025deep,
  title={Deep learning--based precontrast CT parcellation for MRI-free brain amyloid pet quantification},
  author={Choo, Kyobin and Joo, Jaehoon and Lee, Sangwon and Kim, Daesung and Lim, Hyunkeong and Kim, Dongwoo and Kang, Seongjin and Hwang, Seong Jae and Yun, Mijin},
  journal={Clinical Nuclear Medicine},
  volume={50},
  number={5},
  pages={e262--e270},
  year={2025},
  publisher={LWW}
}

@article{matsuda2021amyloid,
  title={Amyloid PET quantification using low-dose CT-guided anatomic standardization},
  author={Matsuda, Hiroshi and Yamao, Tensho and Shakado, Mitsuru and Shigemoto, Yoko and Okita, Kyoji and Sato, Noriko},
  journal={EJNMMI research},
  volume={11},
  number={1},
  pages={125},
  year={2021},
  publisher={Springer}
}

@article{nguyen2020prevalence,
  title={Prevalence and financial impact of claustrophobia, anxiety, patient motion, and other patient events in magnetic resonance imaging},
  author={Nguyen, Xuan V and Tahir, Sana and Bresnahan, Brian W and Andre, Jalal B and Lang, Elvira V and Mossa-Basha, Mahmud and Mayr, Nina A and Bourekas, Eric C},
  journal={Topics in Magnetic Resonance Imaging},
  volume={29},
  number={3},
  pages={125--130},
  year={2020},
  publisher={LWW}
}

@article{landau2023quantification,
  title={Quantification of amyloid beta and tau PET without a structural MRI},
  author={Landau, Susan M and Ward, Tyler J and Murphy, Alice and Iaccarino, Leonardo and Harrison, Theresa M and La Joie, Renaud and Baker, Suzanne and Koeppe, Robert A and Jagust, William J and Alzheimer's Disease Neuroimaging Initiative},
  journal={Alzheimer's \& Dementia},
  volume={19},
  number={2},
  pages={444--455},
  year={2023},
  publisher={Wiley Online Library}
}

@article{lopresti2005simplified,
  title={Simplified quantification of Pittsburgh Compound B amyloid imaging PET studies: a comparative analysis},
  author={Lopresti, Brian J and Klunk, William E and Mathis, Chester A and Hoge, Jessica A and Ziolko, Scott K and Lu, Xueling and Meltzer, Carolyn C and Schimmel, Kurt and Tsopelas, Nicholas D and DeKosky, Steven T and others},
  journal={Journal of Nuclear Medicine},
  volume={46},
  number={12},
  pages={1959--1972},
  year={2005},
  publisher={Society of Nuclear Medicine}
}

@article{lamontagne2019oasis3,
  title   = {{OASIS-3}: Longitudinal Neuroimaging, Clinical, and Cognitive
             Dataset for Normal Aging and {Alzheimer} Disease},
  author  = {LaMontagne, Pamela J. and Benzinger, Tammie L. S. and
             Morris, John C. and Keefe, Sarah and Hornbeck, Russ and
             Xiong, Chengjie and Grant, Elizabeth and Hassenstab, Jason and
             Moulder, Krista and Vlassenko, Andrei G. and Raichle, Marcus E. and
             Cruchaga, Carlos and Marcus, Daniel},
  journal = {medRxiv},
  year    = {2019},
  doi     = {10.1101/2019.12.13.19014902}
}

@article{presotto2018ct,
  title   = {Low-dose {CT} for the spatial normalization of {PET} images:
             A validation procedure for amyloid-{PET} semi-quantification},
  author  = {Presotto, Luca and Iaccarino, Leonardo and Sala, Arianna and
             Vanoli, Emilia Giovanna and Muscio, Cristina and Nigri, Anna and
             Bruzzone, Maria Grazia and Tagliavini, Fabrizio and
             Gianolli, Luigi and Perani, Daniela and Bettinardi, Valentino},
  journal = {NeuroImage: Clinical},
  volume  = {20},
  pages   = {153--160},
  year    = {2018},
  doi     = {10.1016/j.nicl.2018.07.013}
}

@article{ashburner2005unified,
  title   = {Unified segmentation},
  author  = {Ashburner, John and Friston, Karl J.},
  journal = {NeuroImage},
  volume  = {26},
  number  = {3},
  pages   = {839--851},
  year    = {2005},
  doi     = {10.1016/j.neuroimage.2005.02.018}
}

@article{desikan2006automated,
  title   = {An automated labeling system for subdividing the human cerebral
             cortex on {MRI} scans into gyral based regions of interest},
  author  = {Desikan, Rahul S. and S{\'e}gonne, Florent and Fischl, Bruce and
             Quinn, Brian T. and Dickerson, Bradford C. and Blacker, Deborah and
             Buckner, Randy L. and Dale, Anders M. and Maguire, R. Paul and
             Hyman, Bradley T. and Albert, Marilyn S. and Killiany, Ronald J.},
  journal = {NeuroImage},
  volume  = {31},
  number  = {3},
  pages   = {968--980},
  year    = {2006},
  doi     = {10.1016/j.neuroimage.2006.01.021}
}

@article{fischl2012freesurfer,
  title={FreeSurfer},
  author={Fischl, Bruce},
  journal={Neuroimage},
  volume={62},
  number={2},
  pages={774--781},
  year={2012},
  publisher={Elsevier}
}

@article{choi2026gwangju,
  title={The Gwangju Alzheimer's \& Related Dementias (GARD) cohort: Over a decade of Asia's largest longitudinal multimodal study},
  author={Choi, Kyu Yeong and Kang, Sarang and Cook, Seungho and Li, Donghe and Choi, Yu Yong and Seo, Eun Hyun and Han, Xudong and Park, Jung Eun and Lee, Suyeon and Lee, Sunjae and others},
  journal={Alzheimer's \& Dementia},
  volume={22},
  number={1},
  pages={e70981},
  year={2026},
  publisher={Wiley Online Library}
}

@article{su2013quantitative,
  title   = {Quantitative analysis of {PiB-PET} with {FreeSurfer ROIs}},
  author  = {Su, Yi and D'Angelo, Gina M. and Vlassenko, Andrei G. and
             Zhou, Gongfu and Snyder, Abraham Z. and Marcus, Daniel S. and
             Blazey, Tyler M. and Christensen, Jon J. and Vora, Shaney and
             Morris, John C. and Mintun, Mark A. and Benzinger, Tammie L. S.},
  journal = {PLoS ONE},
  volume  = {8},
  number  = {11},
  pages   = {e73377},
  year    = {2013},
  doi     = {10.1371/journal.pone.0073377}
}

@article{klunk2015centiloid,
  title   = {The {Centiloid} Project: Standardizing quantitative amyloid plaque
             estimation by {PET}},
  author  = {Klunk, William E. and Koeppe, Robert A. and Price, Julie C. and
             Benzinger, Tammie L. S. and Devous, Michael D. and
             Jagust, William J. and Johnson, Keith A. and Mathis, Chester A. and
             Minhas, Davneet and Pontecorvo, Michael J. and Rowe, Christopher C.
             and Skovronsky, Daniel M. and Mintun, Mark A.},
  journal = {Alzheimer's \& Dementia},
  volume  = {11},
  number  = {1},
  pages   = {1--15.e4},
  year    = {2015},
  doi     = {10.1016/j.jalz.2014.07.003}
}

@inproceedings{mao2017least,
  title={Least squares generative adversarial networks},
  author={Mao, Xudong and Li, Qing and Xie, Haoran and Lau, Raymond YK and Wang, Zhen and Paul Smolley, Stephen},
  booktitle={Proceedings of the IEEE international conference on computer vision},
  pages={2794--2802},
  year={2017}
}

@inproceedings{kim2024adaptive,
  title={Adaptive latent diffusion model for 3d medical image to image translation: Multi-modal magnetic resonance imaging study},
  author={Kim, Jonghun and Park, Hyunjin},
  booktitle={Proceedings of the IEEE/CVF Winter conference on applications of computer Vision},
  pages={7604--7613},
  year={2024}
}

@inproceedings{choo2024slice,
  title={Slice-consistent 3d volumetric brain ct-to-mri translation with 2d brownian bridge diffusion model},
  author={Choo, Kyobin and Jun, Youngjun and Yun, Mijin and Hwang, Seong Jae},
  booktitle={International Conference on Medical Image Computing and Computer-Assisted Intervention},
  pages={657--667},
  year={2024},
  organization={Springer}
}

@inproceedings{li2023bbdm,
  title={Bbdm: Image-to-image translation with brownian bridge diffusion models},
  author={Li, Bo and Xue, Kaitao and Liu, Bin and Lai, Yu-Kun},
  booktitle={Proceedings of the IEEE/CVF conference on computer vision and pattern Recognition},
  pages={1952--1961},
  year={2023}
}

@article{kim2025automated,
  title={Automated quantification of brain PET in PET/CT using deep learning-based CT-to-MR translation: a feasibility study},
  author={Kim, Daesung and Choo, Kyobin and Lee, Sangwon and Kang, Seongjin and Yun, Mijin and Yang, Jaewon},
  journal={European Journal of Nuclear Medicine and Molecular Imaging},
  volume={52},
  number={8},
  pages={2959--2967},
  year={2025},
  publisher={Springer}
}

@article{edison2013,
  title={Comparison of {MRI} based and {PET} template based approaches in the quantitative analysis of amyloid imaging with {PIB-PET}},
  author={Edison, Paul and Carter, Stephen F. and Rinne, Juha O. and Gelosa, Giorgio and Herholz, Karl and Nordberg, Agneta and Brooks, David J. and Hinz, Rainer},
  journal={NeuroImage},
  volume={70},
  pages={423--433},
  year={2013},
  publisher={Elsevier}
}

@article{lei2018distribution,
  title   = {Distribution-free predictive inference for regression},
  author  = {Lei, Jing and G'Sell, Max and Rinaldo, Alessandro and
             Tibshirani, Ryan J. and Wasserman, Larry},
  journal = {Journal of the American Statistical Association},
  volume  = {113},
  number  = {523},
  pages   = {1094--1111},
  year    = {2018},
  doi     = {10.1080/01621459.2017.1307116}
}

@article{angelopoulos2023gentle,
  title   = {Conformal prediction: A gentle introduction},
  author  = {Angelopoulos, Anastasios N. and Bates, Stephen},
  journal = {Foundations and Trends in Machine Learning},
  volume  = {16},
  number  = {4},
  pages   = {494--591},
  year    = {2023},
  doi     = {10.1561/2200000101}
}

@inproceedings{lakshminarayanan2017simple,
  title     = {Simple and Scalable Predictive Uncertainty Estimation using
               Deep Ensembles},
  author    = {Lakshminarayanan, Balaji and Pritzel, Alexander and
               Blundell, Charles},
  booktitle = {Advances in Neural Information Processing Systems (NeurIPS)},
  volume    = {30},
  pages     = {6402--6413},
  year      = {2017}
}

@article{klunk2004imaging,
  title   = {Imaging brain amyloid in {Alzheimer's} disease with {Pittsburgh}
             Compound-{B}},
  author  = {Klunk, William E. and Engler, Henry and Nordberg, Agneta and
             Wang, Yanming and Blomqvist, Gunnar and Holt, Daniel P. and
             Bergstr{\"o}m, Mats and Savitcheva, Irina and Huang, Guo-Feng and
             Estrada, Sergio and others},
  journal = {Annals of Neurology},
  volume  = {55},
  number  = {3},
  pages   = {306--319},
  year    = {2004},
  doi     = {10.1002/ana.20009}
}

@article{jack2017defining,
  title={Defining imaging biomarker cut points for brain aging and Alzheimer's disease},
  author={Jack Jr, Clifford R and Wiste, Heather J and Weigand, Stephen D and Therneau, Terry M and Lowe, Val J and Knopman, David S and Gunter, Jeffrey L and Senjem, Matthew L and Jones, David T and Kantarci, Kejal and others},
  journal={Alzheimer's \& Dementia},
  volume={13},
  number={3},
  pages={205--216},
  year={2017},
  publisher={Wiley Online Library}
}

@article{tibshirani2019conformal,
  title={Conformal prediction under covariate shift},
  author={Tibshirani, Ryan J and Foygel Barber, Rina and Candes, Emmanuel and Ramdas, Aaditya},
  journal={Advances in neural information processing systems},
  volume={32},
  year={2019}
}

@article{kang2023fast,
  title={Fast and accurate amyloid brain PET quantification without MRI using deep neural networks},
  author={Kang, Seung Kwan and Kim, Daewoon and Shin, Seong A and Kim, Yu Kyeong and Choi, Hongyoon and Lee, Jae Sung},
  journal={Journal of Nuclear Medicine},
  volume={64},
  number={4},
  pages={659--666},
  year={2023},
  publisher={Society of Nuclear Medicine}
}

@article{bailly2015precuneus,
  title={Precuneus and Cingulate Cortex Atrophy and Hypometabolism in Patients with Alzheimer’s Disease and Mild Cognitive Impairment: MRI and 18F-FDG PET Quantitative Analysis Using FreeSurfer},
  author={Bailly, Matthieu and Destrieux, Christophe and Hommet, Caroline and Mondon, Karl and Cottier, Jean-Philippe and Beaufils, Emilie and Vierron, Emilie and Vercouillie, Johnny and Ibazizene, Meziane and Voisin, Thierry and others},
  journal={BioMed research international},
  volume={2015},
  number={1},
  pages={583931},
  year={2015},
  publisher={Wiley Online Library}
}

@article{mishra2017av,
  title={AV-1451 PET imaging of tau pathology in preclinical Alzheimer disease: defining a summary measure},
  author={Mishra, Shruti and Gordon, Brian A and Su, Yi and Christensen, Jon and Friedrichsen, Karl and Jackson, Kelley and Hornbeck, Russ and Balota, David A and Cairns, Nigel J and Morris, John C and others},
  journal={Neuroimage},
  volume={161},
  pages={171--178},
  year={2017},
  publisher={Elsevier}
}

@inproceedings{cohen2018distribution,
  title     = {Distribution matching losses can hallucinate features in medical
               image translation},
  author    = {Cohen, Joseph Paul and Luck, Margaux and Honari, Sina},
  booktitle = {Medical Image Computing and Computer Assisted Intervention
               (MICCAI)},
  series    = {LNCS},
  volume    = {11070},
  pages     = {529--536},
  year      = {2018},
  doi       = {10.1007/978-3-030-00928-1_60}
}

@article{thummerer2023synthrad,
  title   = {{SynthRAD2023} {Grand} {Challenge} dataset: Generating synthetic
             {CT} for radiotherapy},
  author  = {Thummerer, Adrian and van der Bijl, Erik and Galapon, Arthur and
             Verhoeff, Joost J. C. and Langendijk, Johannes A. and Both, Stefan
             and van den Berg, Cornelis A. T. and Maspero, Matteo},
  journal = {Medical Physics},
  volume  = {50},
  number  = {7},
  pages   = {4664--4674},
  year    = {2023},
  doi     = {10.1002/mp.16529}
}

@article{hoopes2022synthstrip,
  title   = {{SynthStrip}: skull-stripping for any brain image},
  author  = {Hoopes, Andrew and Mora, Jocelyn S. and Dalca, Adrian V. and
             Fischl, Bruce and Hoffmann, Malte},
  journal = {NeuroImage},
  volume  = {260},
  pages   = {119474},
  year    = {2022},
  doi     = {10.1016/j.neuroimage.2022.119474}
}

@article{klein2010elastix,
  title   = {elastix: a toolbox for intensity-based medical image registration},
  author  = {Klein, Stefan and Staring, Marius and Murphy, Keelin and
             Viergever, Max A. and Pluim, Josien P. W.},
  journal = {IEEE Transactions on Medical Imaging},
  volume  = {29},
  number  = {1},
  pages   = {196--205},
  year    = {2010},
  doi     = {10.1109/TMI.2009.2035616}
}

@article{grothe2017invivo,
  title   = {In vivo staging of regional amyloid deposition},
  author  = {Grothe, Michel J. and Barthel, Henryk and Sep{\'u}lcre, Jorge and
             Dyrba, Martin and Sabri, Osama and Teipel, Stefan J.},
  journal = {Neurology},
  volume  = {89},
  number  = {20},
  pages   = {2031--2038},
  year    = {2017},
  doi     = {10.1212/WNL.0000000000004643}
}

\appendix
\onecolumn
\renewcommand{\thefigure}{A.\arabic{figure}}
\renewcommand{\thetable}{A.\arabic{table}}
\setcounter{figure}{0}
\setcounter{table}{0}

\section{Supplementary material}
\label{sec:supp}

\subsection{Implementation details}
\label{sec:supp:impl}

Table~\ref{tab:supp:impl} lists the settings needed to situate the results. The complete configuration will be released with the code so that the pipeline can be reproduced.

\begin{table}[!ht]
\centering
\caption{Key implementation settings of the proposed pipeline.}
\label{tab:supp:impl}
\small
\begin{tabular}{@{}p{0.30\textwidth} p{0.63\textwidth}@{}}
\toprule
\textbf{Item} & \textbf{Setting}\\
\midrule
Framework & PyTorch 2.6\\
Optimizer & Adam, lr $10^{-4}$, bfloat16 autocast; EMA decay 0.995\\
Effective batch size & 16 (a mini-batch of 2 with 8-step gradient accumulation on a single A100 GPU)\\
Synthesis input/output & 15 axial CT slices $\rightarrow$ central 3 MRI slices (H3)\\
Diffusion backbone & BBDM, U-Net score network\\
Diffusion process & Brownian bridge with $T=1000$ steps, linear $m_t$, maximum variance 1, the network is trained to predict $m_t(y-x_0)+\sqrt{\delta_t}\,\epsilon$ with an $\ell_1$ loss\\
Training length & Up to 150{,}000 optimizer steps; the checkpoint with the best validation-set PSNR/$\mathrm{SSIM}_{3\mathrm{D}}$ was retained (125{,}000 steps for the final model)\\
Conditioning & Target-MRI intensity histogram + categorical data-source label (H2)\\
Inference & Deterministic ($\eta=0$) sampling with 50 steps and the ISTA-style slice-consistent procedure of Choo et al.\ (one correction step, step size 0.5), using EMA weights; the histogram condition is the mean histogram of the training subjects of the source dataset matching the input CT, and the data-source label is that of the input CT (OASIS-3 for GARD)\\
Ensemble & $M=5$ members: the primary model (seed 1337) and four replicas trained with seeds 42, 123, 456, and 789; identical data, architecture, and hyperparameters, differing only in the random seed (initialization, mini-batch order, and diffusion noise)\\
Segmentation & FreeSurfer \texttt{recon-all} 7.2.0 (\texttt{aparc+aseg})\\
Conformal quantile & finite-sample-corrected $\lceil(n_\text{cal}+1)(1-\alpha)\rceil/n_\text{cal}$, $n_\text{cal}=93$\\
\bottomrule
\end{tabular}
\end{table}

\FloatBarrier
\subsection{CT-to-MRI synthesis baseline implementation details}
\label{sec:supp:baselines}

The implementation settings of the four CT-to-MRI synthesis baselines are
summarized in Table~\ref{tab:supp:baselines}. The following paragraphs
describe how the published methods or released code were adapted to the
present task.

\paragraph{Common protocol}
All four CT-to-MRI synthesis baselines used the paired CT--MRI
preprocessing described in Section~\ref{sec:methods:preprocessing},
including CT-to-T1 registration, brain extraction,
MNI152 alignment, and the same intensity normalization and
$176\times208\times176$ crop.
Inputs were rescaled to
$[-1,1]$ for model training. Each baseline was trained separately on the
OASIS-3 ($N=773$) and Severance ($N=180$) training sets; the OASIS-3
models were also applied to GARD without retraining. A fixed random
seed of 1337 was used for the baseline data splits and initialization.
All methods were evaluated using the same brain-masked PSNR and
$\mathrm{SSIM}_{3\mathrm{D}}$ metrics, FreeSurfer parcellation, and PET
quantification procedures.

\begin{table}[!ht]
\centering
\caption{Implementation settings of the four CT-to-MRI synthesis
baselines in this study. Training length and checkpoint selection are
reported separately for OASIS-3 and Severance where applicable.
EMA: exponential moving average; ISTA: iterative
shrinkage-thresholding-style sampling.}
\label{tab:supp:baselines}
\footnotesize
\setlength{\tabcolsep}{3pt}
\renewcommand{\arraystretch}{1.08}
\begin{tabular*}{\textwidth}{@{\extracolsep{\fill}}
>{\raggedright\arraybackslash}p{0.14\textwidth}
>{\raggedright\arraybackslash}p{0.21\textwidth}
>{\raggedright\arraybackslash}p{0.21\textwidth}
>{\raggedright\arraybackslash}p{0.17\textwidth}
>{\raggedright\arraybackslash}p{0.19\textwidth}@{}}
\toprule
\textbf{Setting}
& \textbf{Choo2024}
& \textbf{ALDM}
& \textbf{Kim2025}
& \textbf{LSGAN} \\
\midrule
Implementation source
& Released CT2MRI code [\ref{sref:choo2024slice}]
& Released network modules [\ref{sref:kim2024adaptive}]
& Re-implemented from the publication [\ref{sref:kim2025automated}]
& 3D pix2pix with least-squares loss [\ref{sref:isola2017imagetoimage},\ref{sref:mao2017least}] \\

Input $\rightarrow$ output
& 3 CT slices $\rightarrow$ 3 MRI slices
& 3D CT $\rightarrow$ 3D MRI
& 3 CT slices $\rightarrow$ central MRI slice
& 3D CT $\rightarrow$ 3D MRI \\

Network
& 2D BBDM U-Net; MRI histogram condition
& 3D first-stage encoder/decoder with MS-SPADE; latent diffusion U-Net
& 2.5D residual U-Net (16/32/64 encoder channels; six residual blocks)
& 3D U-Net generator and 3D PatchGAN discriminator \\

Training objective
& Brownian bridge, $\ell_1$; $T=1000$
& First stage $\ell_1$; second-stage $\epsilon$-prediction MSE, $T=1000$
& $\ell_2$
& Least-squares adversarial $+100\ell_1$ \\

Optimizer
& Adam, $10^{-4}$, $\beta=(0.9,0.999)$; EMA 0.995
& AdamW, $10^{-4}$, weight decay $10^{-4}$; EMA 0.9999
& Adam, $10^{-4}$, $\beta=(0.5,0.999)$
& Adam, $2\times10^{-4}$, $\beta=(0.5,0.999)$ \\

Effective batch
& 16 (2 $\times$ 8 accumulation)
& 4 (1 $\times$ 4 accumulation)
& 16
& 4 (1 $\times$ 4 accumulation) \\

Training length
& 100{,}000 optimizer steps
& 500 / 2{,}200 epochs ($\approx$96.5k / 99k optimizer steps)
& Up to 300 epochs
& 200 epochs \\

Checkpoint
& Best validation performance
& Best validation performance
& Validation-selected; epoch 15 / 20
& Lowest validation MAE, assessed every 20 epochs \\

Inference
& 50-step ISTA, $\eta=0$; training-set mean MRI histogram
& 100-step DDIM, $\eta=0$
& Independent slice prediction
& Deterministic \\

Precision
& bfloat16 autocast
& bfloat16 autocast
& bfloat16 autocast
& FP32 \\
\bottomrule
\end{tabular*}
\end{table}

\paragraph{Choo2024 (slice-consistent BBDM)}
We adapted the authors' released implementation
(\url{https://github.com/MICV-yonsei/CT2MRI})
[\ref{sref:choo2024slice}], retaining its three-slice architecture,
histogram conditioning, and slice-consistent ISTA-style sampling.
The released $160\times160$ slices were replaced by the
$208\times176$ slices from our MNI-space preprocessing. Training used
the released optimizer settings, an effective batch size of 16, and
100{,}000 optimizer steps. At inference, the MRI histogram condition
was the training-set mean rather than a histogram from the test
subject's MRI. We used bfloat16 autocast, clipped the predicted clean
image to $[-1,1]$ during sampling, and selected the checkpoint on
the validation set rather than using the final checkpoint.

\paragraph{ALDM (adaptive latent diffusion model)}
We used the network modules from the authors' released code
(\url{https://github.com/jongdory/ALDM})
[\ref{sref:kim2024adaptive}] and wrote a PyTorch training loop for
paired CT-to-T1 synthesis. The first stage followed the released
VQ-GAN architecture with an $\ell_1$ training objective; its
discriminator and codebook loss weights were set to zero. It was
trained on CT--T1 pairs and then frozen. We used a single MS-SPADE
target branch and disabled modality-label conditioning because the
task has one target modality. Compared with the publication, our
implementation used a higher learning rate ($10^{-4}$ rather than
$2\times10^{-6}$), a plain linear rather than scaled-linear noise
schedule, and an effective batch size of four on a single 40\,GB
A100. The latent diffusion stage was trained for approximately
100{,}000 optimizer steps per dataset, and inference used
deterministic 100-step DDIM sampling rather than the released code's
200-step stochastic sampling.

\paragraph{Kim2025 (CT-to-MRI U-Net for PET quantification)}
No public implementation was available for Kim et al.\
[\ref{sref:kim2025automated}]. We re-implemented the CT-to-MRI
translation network using the architecture depicted in its
Supplementary Fig.~1: three encoder and decoder blocks, six
residual blocks, and a linear output layer. Where the publication
did not specify layer details, we used bias-free convolutions before
instance normalization. Training followed the reported $\ell_2$
objective and Adam settings. Instead of a fixed 250 epochs, we
trained for up to 300 epochs and selected validation checkpoints
at epoch 15 on OASIS-3 and epoch 20 on Severance. At inference,
independently predicted slices were stacked into a volume. For all
synthesis methods, including this baseline, the generated image was
parcellated with FreeSurfer rather than the separate segmentation
network of Kim et al.

\paragraph{LSGAN (3D conditional GAN with least-squares loss)}
This baseline extended the pix2pix generator and discriminator to 3D
[\ref{sref:isola2017imagetoimage}], using the least-squares
adversarial objective [\ref{sref:mao2017least}] and an $\ell_1$
reconstruction weight of 100. The generator was a 3D U-Net and the
discriminator a 3D PatchGAN, each with a base width of 64 channels.
The official pix2pix implementation supplied the layer and loss
definitions
(\url{https://github.com/junyanz/pytorch-CycleGAN-and-pix2pix}).
Paired CT--MRI volumes were augmented with TorchIO flips
($p=0.5$) and affine transforms (scaling $\pm5\%$, rotation
$\pm10^\circ$, translation $\pm5$ voxels). The model was trained
for 200 epochs; the checkpoint with the lowest validation MAE
within the brain bounding box, assessed every 20 epochs, was
retained. Unlike the pix2pix defaults, we did not apply generator
dropout, explicit Gaussian weight initialization, or learning-rate
decay.

\FloatBarrier

\subsection{Non-synthesis MRI-free baseline implementation details}
\label{sec:supp:nonsynthesis}

\paragraph{Common protocol}
We evaluated two PET-only and two CT-based implementations on the same OASIS-3 test set, using the same static PiB images as the real MRI-based reference. Processing used the SPM12 batch modules in the SPM25 standalone release. Anatomical target-region definitions matched the reference pipeline. Atlas and reference-region sources are listed in Table~\ref{tab:supp:nonsynthesis_settings}. The OASIS-derived and CT-based implementations used a 1-mm isotropic grid with a bounding box from $(-90,-126,-72)$ to $(90,90,108)$~mm in MNI152 space. The GAAIN-derived implementation used the SPM default bounding box and a 2-mm isotropic grid.

The GAAIN-derived implementation recentered the image origin and rigidly coregistered PET to the template before normalization. The OASIS-derived implementation used the intensity-weighted image center for its initial translation. No separate initial alignment was applied in Presotto2018. Matsuda2021 used the skull-based alignment described below.

\begin{table}[!ht]
\centering
\caption{Main differences between the non-synthesis implementations. Both PET-only implementations use SPM Old Normalise. Both CT-based implementations apply CT-derived unified-segmentation deformations to PET. DK denotes Desikan--Killiany, and GM denotes grey matter.}
\label{tab:supp:nonsynthesis_settings}
\footnotesize
\setlength{\tabcolsep}{4pt}
\renewcommand{\arraystretch}{1.15}
\begin{tabular*}{\textwidth}{@{\extracolsep{\fill}}
>{\raggedright\arraybackslash}p{0.21\textwidth}
>{\raggedright\arraybackslash}p{0.44\textwidth}
>{\raggedright\arraybackslash}p{0.29\textwidth}@{}}
\toprule
\textbf{Implementation} & \textbf{Template or normalization settings} & \textbf{Target atlas / reference region} \\
\midrule
\multicolumn{3}{@{}l}{\textit{PET-only template-based normalization}} \\
PET-template (GAAIN-derived) & PiB template constructed from the public GAAIN AD-100 and YC-0 calibration PET/MRI datasets & FreeSurfer DK parcellation of the MNI152 template / GAAIN cerebellar GM \\
PET-template (OASIS-derived) & PiB template from 197 OASIS-3 subjects. Normalization used 8-mm source smoothing and default regularization. & Population atlas from 200 OASIS-3 FreeSurfer parcellations / cerebellar cortex \\
\midrule
\multicolumn{3}{@{}l}{\textit{CT-based anatomical normalization}} \\
Presotto2018 & HU $<-300$ set to $-1024$. Bias correction disabled. Gaussians per tissue class were 1/1/2/2/2/2. & DK atlas from atlasreader (cvs\_avg35 in MNI152) / cerebellar cortex \\
Matsuda2021 & Skull-based initialization and CT coregistration to avg152T1. Default segmentation settings without HU thresholding. & DK atlas from atlasreader (cvs\_avg35 in MNI152) / cerebellar cortex \\
\bottomrule
\end{tabular*}
\end{table}

\paragraph{PET-only implementations}
These implementations adapted the template-based approaches of Edison et al.\ [\ref{sref:edison2013}] and Landau et al.\ [\ref{sref:landau2023quantification}]. Both templates were averages of PiB images spatially normalized using the corresponding MRIs. Evaluation required PET only. The GAAIN-derived template used the publicly available AD-100 and YC-0 PiB calibration datasets [\ref{sref:klunk2015centiloid}]. The public release comprises 45 Alzheimer's disease patients and 34 young controls and provides PET images with corresponding MRIs. For template construction, each PET was rigidly coregistered to its MRI before MRI-based spatial normalization. PET images were scaled by the mean uptake in the cerebellar GM reference region before averaging. The OASIS-derived template used cerebellar-cortex-scaled PET from 197 subjects (47 A+/CI, 49 A+/CU, 16 A$-$/CI, and 85 A$-$/CU), where A+ denotes Centiloid $>20$ and CI denotes CDR $>0$. Its atlas used 200 separate cognitively unimpaired subjects. OASIS-3 template and atlas subjects were excluded from the validation, calibration, and test sets. Atlas construction involved warping binary FreeSurfer masks to MNI152 space, smoothing at 1.5-mm FWHM, averaging and scaling each regional map to $[0,1]$, and assigning the highest-ranking label to each voxel. The two pipelines differ in template, atlas, and reference definition, rather than template source alone.

\paragraph{CT-based implementations}
In both implementations, PET was rigidly coregistered to the same-day CT. CT was normalized to MNI152 space using SPM unified segmentation [\ref{sref:ashburner2005unified}]. The resulting deformation was applied to PET. The implementations adapted the preprocessing and segmentation approaches of Presotto et al.\ [\ref{sref:presotto2018ct}] and Matsuda et al.\ [\ref{sref:matsuda2021amyloid}] to PiB and the present target/reference definitions. For Matsuda2021, we added automatic orientation. The largest skull component above 300~HU within 170~mm of the vertex was aligned to the SPM bone probability map using rigid alignment with isotropic scaling and multiple starting rotations. PET and CT received the same initialization transform. Unspecified settings used the SPM defaults.

\FloatBarrier
\subsection{Segmentation-teacher feedback}
\label{sec:supp:teacher}

As an additional anatomical regularization strategy, we explored a
segmentation-teacher module that provided an auxiliary training signal to the
CT-to-MRI synthesis model. The motivation was to encourage the generated MRI
to preserve tissue boundaries and anatomical structures that may not be fully
captured by the original image-level synthesis objective alone.

The teacher was a pretrained 2.5D depthwise-separable U-Net with approximately
2.2 million parameters. It received three adjacent T1-weighted MRI slices as
input and predicted seven tissue classes, including background. The model was
pretrained using FreeSurfer-derived segmentation labels from 773 OASIS-3
training pairs (620 subjects). Gaussian intensity noise and spatial augmentation were
applied during teacher pretraining to improve robustness to variation in image
appearance.

During CT-to-MRI training, the pretrained teacher network was frozen and used
only to provide an auxiliary anatomical loss. Both the generated MRI and the
corresponding real MRI were passed through the teacher network, and their
predicted tissue probabilities were compared using a Dice-based loss. The loss
was computed within the brain mask and averaged over the six foreground tissue
classes:

\begin{equation}
\mathcal{L}_{\mathrm{seg}}
=
1 -
\frac{1}{C}
\sum_{c=1}^{C}
\frac{
2 \sum_{v} p_{v,c}^{\mathrm{syn}} p_{v,c}^{\mathrm{real}} + \epsilon
}{
\sum_{v} p_{v,c}^{\mathrm{syn}}
+
\sum_{v} p_{v,c}^{\mathrm{real}}
+
\epsilon
},
\end{equation}

where $C=6$, $p_{v,c}^{\mathrm{syn}}$ and
$p_{v,c}^{\mathrm{real}}$ denote the teacher-predicted tissue probabilities
for class $c$ at voxel $v$ from the synthetic and real MRI, respectively, and
$\epsilon$ is a small constant for numerical stability.

The segmentation-teacher loss was added to the original synthesis objective as

\begin{equation}
\mathcal{L}_{\mathrm{total}}
=
\mathcal{L}_{\mathrm{BBDM}}
+
\lambda_{\mathrm{seg}} \mathcal{L}_{\mathrm{seg}},
\end{equation}

where $\mathcal{L}_{\mathrm{BBDM}}$ denotes the original CT-to-MRI synthesis
objective and $\lambda_{\mathrm{seg}}$ controls the contribution of the
segmentation-teacher regularization. We selected $\lambda_{\mathrm{seg}}$
based on validation-set performance by evaluating
$\{1,\,0.1,\,0.03,\,0.01,\,0.003,\,0.001,\,3\times10^{-4},\,10^{-4}\}$.
The final setting used $\lambda_{\mathrm{seg}}=0.001$.

The teacher network was used only during training and introduced no additional
input or computation at inference. In the cumulative ablation analysis,
segmentation-teacher feedback (H4) did not provide a statistically significant
additional improvement beyond H1--H3 and was therefore excluded from the final
model.

\FloatBarrier
\subsection{Ablation study of the proposed design choices}
\label{sec:supp:ablation-sec}

We provide the detailed results of the cumulative ablation analysis summarized in Section~\ref{sec:design_results}. Starting from the Choo2024 BBDM backbone, the proposed design choices were introduced sequentially as
$\text{BBDM} \rightarrow +\text{H1} \rightarrow +\text{H2} \rightarrow +\text{H3} \rightarrow +\text{H4}$.
H1 denotes multi-dataset training, H2 data-source conditioning, H3 expanded multi-slice anatomical context using the 15-slice input selected in Section~\ref{sec:supp:context}, and H4 segmentation-teacher feedback. All configurations were evaluated on held-out validation sets from OASIS-3 ($N=20$) and Severance ($N=10$), which are disjoint from the test cohorts used for the main synthesis evaluation. PSNR and $\mathrm{SSIM}_{3\mathrm{D}}$ were computed within the brain mask.

Because every subject was evaluated using all model configurations, comparisons were performed in a paired manner. Each newly added component was compared with the immediately preceding configuration using a one-sided Wilcoxon signed-rank test. The reported significance therefore reflects the marginal effect of adding each component to the cumulative model rather than its standalone effect. One-sided tests were used because the component ablation evaluates prespecified directional hypotheses that each added design choice improves synthesis performance.

Detailed results are reported in Table~\ref{tab:supp:ablation}. H1 and H2 provide smaller incremental improvements within the cumulative sequence, although the H1 improvement in Severance PSNR does not reach statistical significance. H3 provides the largest marginal performance gain among the evaluated design choices. H4 provides no significant additional improvement in either cohort or metric and is excluded from the final model.

\begin{table}[!ht]
\centering
\caption{Cumulative ablation analysis on the validation sets
(Severance $N=10$ and OASIS-3 $N=20$).
Results are reported as mean $\pm$ SD for masked PSNR and
$\mathrm{SSIM}_{3\mathrm{D}}$.
Superscripts indicate one-sided paired Wilcoxon significance relative to
the preceding configuration
($^{*}p<0.05$, $^{**}p<0.01$, $^{***}p<0.001$,
$^\text{ns}$ not significant).}
\label{tab:supp:ablation}
\small
\begin{tabular}{lcccc}
\toprule
& \multicolumn{2}{c}{\textbf{Severance} ($N=10$)} & \multicolumn{2}{c}{\textbf{OASIS-3} ($N=20$)}\\
\cmidrule(lr){2-3}\cmidrule(lr){4-5}
\textbf{Model} & PSNR $\uparrow$ & $\mathrm{SSIM}_{3\mathrm{D}}$ $\uparrow$ & PSNR $\uparrow$ & $\mathrm{SSIM}_{3\mathrm{D}}$ $\uparrow$\\
\midrule
BBDM~[\ref{sref:choo2024slice}] & $25.26 \pm 1.77$ & $0.912 \pm 0.034$ & $23.51 \pm 0.48$ & $0.846 \pm 0.016$\\
$+$H1 & $25.46 \pm 1.78^\text{ns}$ & $0.919 \pm 0.032^{***}$ & $23.59 \pm 0.48^{*}$ & $0.848 \pm 0.016^{*}$\\
$+$H1$+$H2 & $25.64 \pm 1.86^{**}$ & $0.922 \pm 0.032^{***}$ & $23.69 \pm 0.50^{*}$ & $0.852 \pm 0.015^{***}$\\
$+$H1$+$H2$+$H3 \textbf{(proposed)} & $\mathbf{26.26 \pm 1.99}^{**}$ & $\mathbf{0.931 \pm 0.034}^{**}$ & $\mathbf{24.76 \pm 0.53}^{***}$ & $\mathbf{0.881 \pm 0.013}^{***}$\\
\addlinespace[1pt]
\quad $+$H4 \emph{(explored)} & $26.16 \pm 2.08^\text{ns}$ & $0.929 \pm 0.037^\text{ns}$ & $24.37 \pm 0.60^\text{ns}$ & $0.872 \pm 0.014^\text{ns}$\\
\bottomrule
\end{tabular}
\end{table}

\FloatBarrier
\subsection{Effect of multi-slice CT context}
\label{sec:supp:context}

To determine the extent of through-plane CT context used in H3, we evaluated
progressively wider input contexts from the original three-slice configuration
up to 15 consecutive axial CT slices within a practical computational budget.
All other model components were kept fixed across context sizes. Performance
was evaluated on the same held-out validation sets used for the cumulative
ablation analysis.

\begin{figure}[!ht]
    \centering
    \includegraphics[width=\linewidth]{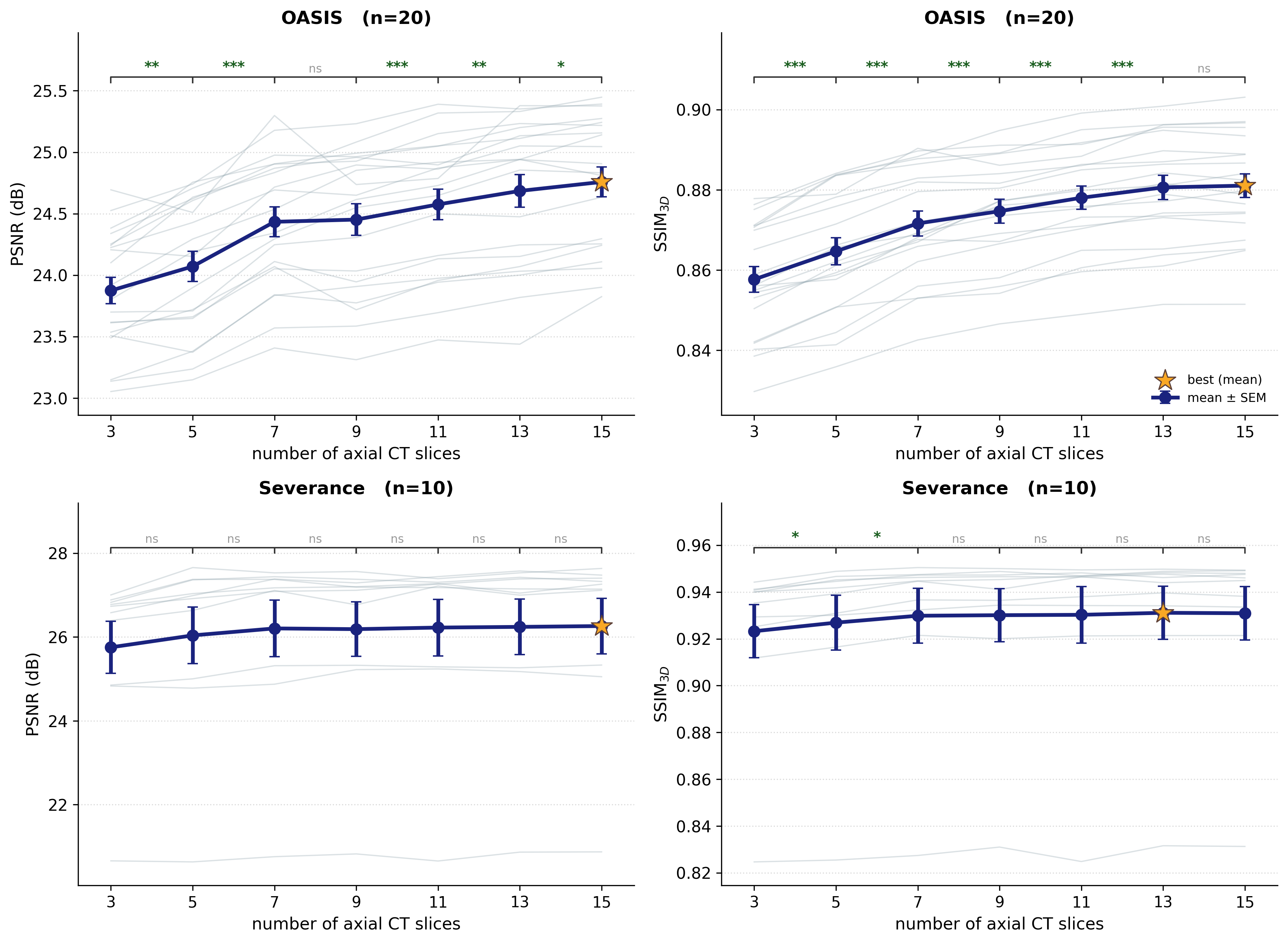}
    \caption{Effect of through-plane CT context on synthesis performance on the validation sets. The number of input axial CT slices was varied from 3 to 15 while all other model components were fixed. Thin lines show paired subject-level performance, and the dark line with error bars denotes the mean $\pm$ SEM. Statistical significance was assessed between consecutive context sizes using two-sided Wilcoxon signed-rank tests with Holm correction for multiple comparisons ($^{*}p<0.05$, $^{**}p<0.01$, $^{***}p<0.001$, $^\mathrm{ns}$ not significant).}
    \label{fig:supp:context}
\end{figure}

On OASIS-3, synthesis fidelity generally improved as the CT context was
expanded, with both PSNR and $\mathrm{SSIM}_{3\mathrm{D}}$ increasing over the
three-slice baseline (Fig.~\ref{fig:supp:context}). On Severance, most of the
improvement was obtained with a smaller increase in context, after which
performance approached a plateau. Given the continued improvement on OASIS-3
and the plateauing performance on Severance, we selected 15 input CT slices
for the final model among the evaluated settings.

Figure~\ref{fig:supp:context} shows the paired subject-level trajectories across all evaluated context sizes. Statistical comparisons between consecutive context sizes were performed using two-sided Wilcoxon signed-rank tests with Holm correction for multiple comparisons.

\FloatBarrier
\subsection{Per-subject paired comparison against Choo2024}
\label{sec:supp:slope}

\begin{figure}[!ht]
\centering
\includegraphics[width=\linewidth]{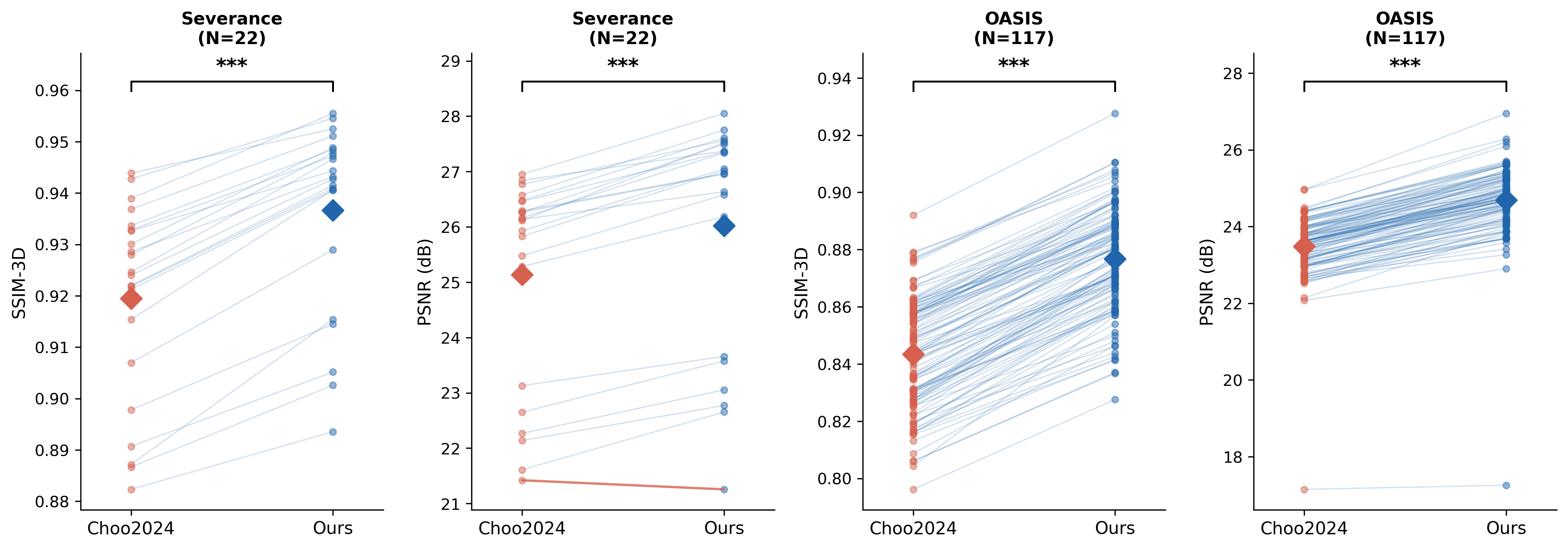}
\caption{Per-subject paired comparison between the proposed method and Choo2024, the strongest baseline, across both cohorts and evaluation metrics. Each thin line connects paired values from a single subject. Blue lines indicate better performance of the proposed method, whereas red lines indicate the opposite. Diamonds denote group means. Statistical significance was assessed using one-sided Wilcoxon signed-rank tests on the paired subject-level differences ($^{***}p < 0.001$).}
\label{fig:supp:slope}
\end{figure}

Figure~\ref{fig:supp:slope} provides the subject-level paired comparison between the proposed method and Choo2024, the strongest baseline, for both cohorts and evaluation metrics. The proposed method outperforms Choo2024 for all subjects in $\mathrm{SSIM}_{3\mathrm{D}}$ in both cohorts and in PSNR on OASIS-3. For Severance PSNR, it performs better in 21 of 22 subjects. All four one-sided Wilcoxon signed-rank tests on the paired subject-level differences are significant at $p<0.001$.

\FloatBarrier
\subsection{Regional SUVR performance of non-synthesis MRI-free baselines}
Regional SUVRs of the non-synthesis pipelines were computed for the six
regions reported in Table~\ref{tab:regional_mae}, using the same regional
definitions and the atlas and reference region of each pipeline
(Table~\ref{tab:supp:nonsynthesis_settings}).

\begin{table}[!ht]
\centering
\caption{MAE of regional SUVR estimates for the non-synthesis MRI-free
baselines ($N=87$), reported as mean $\pm$ SD in SUVR units. The proposed
method is shown for reference using the results from
Table~\ref{tab:regional_mae}. Target regions of the non-synthesis pipelines
were defined in MNI152 space and therefore do not coincide exactly with the
native-space FreeSurfer regions of the reference.}
\label{tab:supp:nonsynthesis_regional}
\footnotesize
\setlength{\tabcolsep}{3pt}
\begin{tabular}{lcccccc}
\toprule
\textbf{Pipeline}
& \textbf{Frontal}
& \textbf{Parietal}
& \makecell{\textbf{Lat.}\\\textbf{temporal}}
& \textbf{Precuneus}
& \makecell{\textbf{Ant.}\\\textbf{cingulate}}
& \makecell{\textbf{Post.}\\\textbf{cingulate}} \\
\midrule
PET-template (GAAIN-derived)
& $0.173 \pm 0.103$
& $0.160 \pm 0.096$
& $0.157 \pm 0.094$
& $0.151 \pm 0.101$
& $0.251 \pm 0.149$
& $0.170 \pm 0.116$ \\

PET-template (OASIS-derived)
& $0.140 \pm 0.082$
& $0.146 \pm 0.073$
& $0.132 \pm 0.074$
& $0.122 \pm 0.075$
& $0.154 \pm 0.105$
& $0.140 \pm 0.099$ \\

Presotto2018
& $0.098 \pm 0.075$
& $0.142 \pm 0.082$
& $0.041 \pm 0.041$
& $0.072 \pm 0.072$
& $0.139 \pm 0.104$
& $0.128 \pm 0.126$ \\

Matsuda2021
& $0.087 \pm 0.082$
& $0.154 \pm 0.084$
& $0.052 \pm 0.050$
& $0.090 \pm 0.081$
& $0.191 \pm 0.122$
& $0.168 \pm 0.133$ \\
\midrule
\textbf{Ours}
& $\mathbf{0.017 \pm 0.015}$
& $\mathbf{0.016 \pm 0.015}$
& $\mathbf{0.017 \pm 0.017}$
& $\mathbf{0.021 \pm 0.023}$
& $\mathbf{0.026 \pm 0.020}$
& $\mathbf{0.023 \pm 0.021}$ \\
\bottomrule
\end{tabular}
\end{table}

\FloatBarrier
\subsection{Regional conformal prediction across target coverage levels}
\label{sec:supp:regional_conformal}

We further evaluated regional SUVR conformal prediction across target
coverage levels of 80\%, 90\%, and 95\%
(Table~\ref{tab:supp:regional_conformal}). Across all six regions,
prediction intervals widened as the target coverage increased. Empirical
coverage generally varied around the corresponding target levels, although
the degree of under- or overcoverage differed across regions.

\begin{figure*}[!ht]
    \centering
    \includegraphics[width=\linewidth]{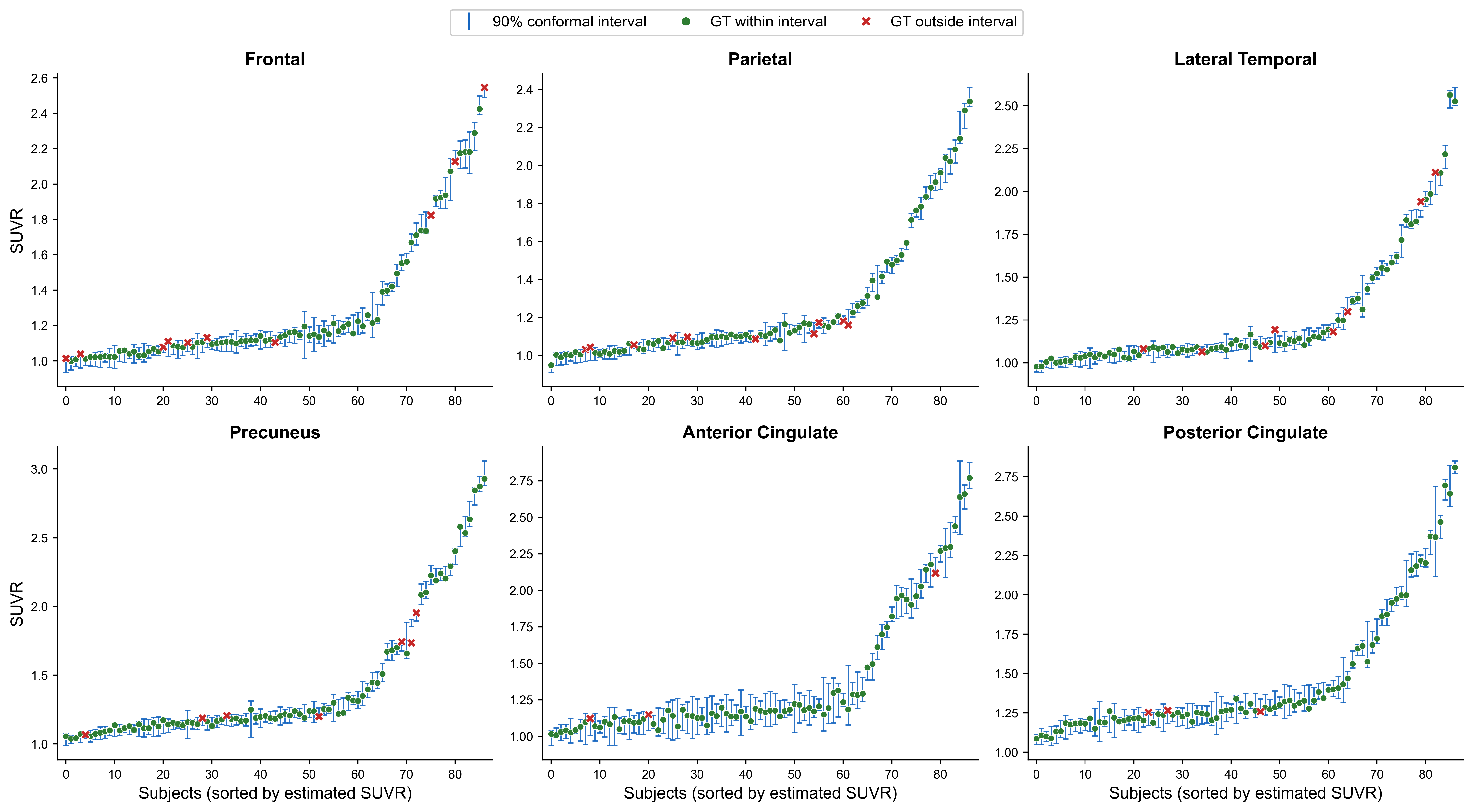}
    \caption{
    Subject-level conformal prediction intervals for the six regional SUVR
    measures at the 90\% target coverage. Within each region, subjects are
    ordered by the ensemble-mean MRI-free SUVR estimate. Green circles and
    red crosses indicate whether the corresponding MRI-based reference SUVR
    falls inside or outside the conformal prediction interval, respectively.
    Empirical coverage and mean prediction interval width for each region,
    including results at the 80\% and 95\% target coverage levels, are
    reported in Table~\ref{tab:supp:regional_conformal}.
    }
    \label{fig:supp:regional_conformal}
\end{figure*}

At the 80\% target, empirical coverage ranged from 77.0\% to 86.2\%.
At the 90\% target, coverage ranged from 88.5\% to 96.6\%, as reported
in the main analysis. Figure~\ref{fig:supp:regional_conformal} shows the
corresponding subject-level prediction intervals at this 90\% target coverage. At the 95\% target, empirical coverage ranged from 89.7\% in the
parietal cortex to 100.0\% in the posterior cingulate cortex. Mean
prediction interval width increased with the target coverage in every region.

\begin{table}[!ht]
\centering
\caption{
Regional conformal prediction performance across target coverage levels.
Empirical coverage and prediction interval width are reported for each regional
SUVR at target coverage levels of 80\%, 90\%, and 95\%.
Empirical coverage denotes the proportion of test subjects whose real
MRI-based SUVR falls within the corresponding prediction interval,
with counts shown in parentheses as covered subjects/total subjects.
Two-sided 95\% Wilson confidence intervals for coverage are shown in
brackets below each coverage value, in percentage units.
Interval width is reported in SUVR units as mean $\pm$ SD across test
subjects ($N=87$).
}
\label{tab:supp:regional_conformal}

\small
\setlength{\tabcolsep}{2pt}
\renewcommand{\arraystretch}{1.05}
\setlength{\medmuskip}{2mu}

\begin{tabular*}{\linewidth}
{@{\extracolsep{\fill}}lcccccc@{}}
\toprule
& \multicolumn{2}{c}{\textbf{80\% target}}
& \multicolumn{2}{c}{\textbf{90\% target}}
& \multicolumn{2}{c}{\textbf{95\% target}} \\
\cmidrule(lr){2-3}
\cmidrule(lr){4-5}
\cmidrule(lr){6-7}

\textbf{Region}
& \textbf{Coverage}
& \textbf{Width}
& \textbf{Coverage}
& \textbf{Width}
& \textbf{Coverage}
& \textbf{Width} \\
\midrule

Frontal
& \makecell[c]{77.0\% (67/87)\\[-1pt]
  {\scriptsize [67.1, 84.6]}}
& $0.070 \pm 0.037$
& \makecell[c]{88.5\% (77/87)\\[-1pt]
  {\scriptsize [80.1, 93.6]}}
& $0.094 \pm 0.049$
& \makecell[c]{95.4\% (83/87)\\[-1pt]
  {\scriptsize [88.8, 98.2]}}
& $0.120 \pm 0.063$ \\
\addlinespace[3pt]

Parietal
& \makecell[c]{83.9\% (73/87)\\[-1pt]
  {\scriptsize [74.8, 90.2]}}
& $0.061 \pm 0.032$
& \makecell[c]{88.5\% (77/87)\\[-1pt]
  {\scriptsize [80.1, 93.6]}}
& $0.069 \pm 0.036$
& \makecell[c]{89.7\% (78/87)\\[-1pt]
  {\scriptsize [81.5, 94.5]}}
& $0.074 \pm 0.039$ \\
\addlinespace[3pt]

\makecell[l]{Lateral\\temporal}
& \makecell[c]{77.0\% (67/87)\\[-1pt]
  {\scriptsize [67.1, 84.6]}}
& $0.056 \pm 0.029$
& \makecell[c]{90.8\% (79/87)\\[-1pt]
  {\scriptsize [82.9, 95.3]}}
& $0.074 \pm 0.039$
& \makecell[c]{93.1\% (81/87)\\[-1pt]
  {\scriptsize [85.8, 96.8]}}
& $0.086 \pm 0.045$ \\
\addlinespace[3pt]

Precuneus
& \makecell[c]{85.1\% (74/87)\\[-1pt]
  {\scriptsize [76.1, 91.1]}}
& $0.075 \pm 0.035$
& \makecell[c]{92.0\% (80/87)\\[-1pt]
  {\scriptsize [84.3, 96.0]}}
& $0.101 \pm 0.046$
& \makecell[c]{94.3\% (82/87)\\[-1pt]
  {\scriptsize [87.2, 97.5]}}
& $0.116 \pm 0.053$ \\
\addlinespace[3pt]

\makecell[l]{Anterior\\cingulate}
& \makecell[c]{82.8\% (72/87)\\[-1pt]
  {\scriptsize [73.5, 89.3]}}
& $0.094 \pm 0.042$
& \makecell[c]{96.6\% (84/87)\\[-1pt]
  {\scriptsize [90.3, 98.8]}}
& $0.185 \pm 0.082$
& \makecell[c]{96.6\% (84/87)\\[-1pt]
  {\scriptsize [90.3, 98.8]}}
& $0.193 \pm 0.086$ \\
\addlinespace[3pt]

\makecell[l]{Posterior\\cingulate}
& \makecell[c]{86.2\% (75/87)\\[-1pt]
  {\scriptsize [77.4, 91.9]}}
& $0.088 \pm 0.050$
& \makecell[c]{96.6\% (84/87)\\[-1pt]
  {\scriptsize [90.3, 98.8]}}
& $0.132 \pm 0.075$
& \makecell[c]{100.0\% (87/87)\\[-1pt]
  {\scriptsize [95.8, 100.0]}}
& $0.180 \pm 0.102$ \\

\bottomrule
\end{tabular*}
\end{table}

\FloatBarrier
\subsection{Sensitivity of amyloid-positivity decision reliability to the MCSUVR cut-off}
\label{sec:supp:mcsuvr_threshold}

At the prespecified MCSUVR cut-off of 1.42, the amyloid-positivity decisions
obtained from our synthetic-MRI-based MRI-free pipeline reproduced the real
MRI-based decisions for all 87 test subjects. We therefore examined whether
this agreement was sensitive to the choice of MCSUVR cut-off and whether
conformal prediction could identify unreliable decisions when disagreements
emerged. The cut-off was varied from 1.30 to 1.50 in increments of 0.01,
while the 90\% conformal prediction intervals were held fixed. At each
cut-off, MRI-free point-estimate decisions were compared with the
corresponding real MRI-based reference decisions, and cases whose prediction
intervals spanned the cut-off were designated indeterminate.

\begin{figure}[!ht]
    \centering
    \includegraphics[width=\columnwidth]
    {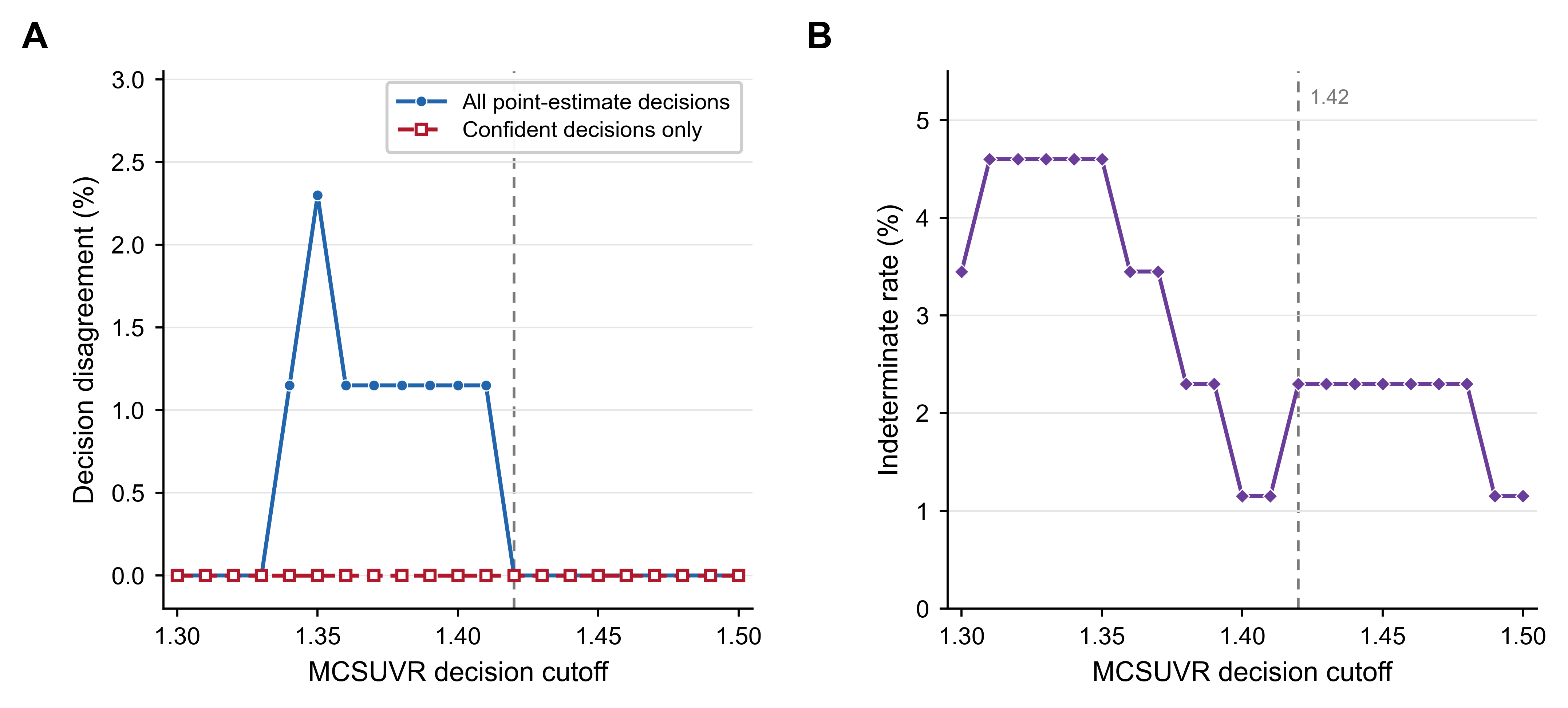}
    \caption{
    Sensitivity of amyloid-positivity decision reliability to the MCSUVR
    cut-off using fixed 90\% conformal prediction intervals.
    The MCSUVR cut-off was varied from 1.30 to 1.50 while the conformal
    prediction intervals were held fixed.
    (A) Disagreement rate between the MRI-free point-estimate
    amyloid-positivity decision and the real MRI-based reference decision
    for all subjects and for subjects retaining a confident conformal
    decision.
    (B) Proportion of subjects designated indeterminate because their
    conformal prediction interval spanned the decision cut-off.
    The vertical dashed line marks the prespecified PiB PET MCSUVR cut-off
    of 1.42 used in the main analysis.
    }
    \label{fig:supp:mcsuvr_threshold}
\end{figure}

Point-estimate disagreement with the real MRI-based reference decision was
rare and occurred only at cut-offs from 1.34 to 1.41
(Fig.~\ref{fig:supp:mcsuvr_threshold}). The maximum disagreement involved
two subjects at a cut-off of 1.35. At every cut-off at which a disagreement
occurred, all discordant point-estimate decisions were designated
indeterminate. This shows that cases in which the MRI-free decision differed
from the real MRI-based reference were consistently flagged by the conformal
decision rule. Conversely, no discordant case remained among subjects
receiving a confident decision, and all confident decisions matched the real
MRI-based reference decision across the evaluated cut-off range.

The proportion of subjects designated indeterminate remained low throughout
the sweep, ranging from 1.1\% to 4.6\% (1--4 of 87 subjects). Accordingly,
83--86 subjects retained a confident decision depending on the cut-off.
Representative cut-offs are summarized in
Table~\ref{tab:supp:mcsuvr_threshold}. Within this test set and evaluated
cut-off range, all observed point-estimate disagreements were therefore
deferred as indeterminate, while decisions retained as confident remained
concordant with the real MRI-based reference decision.

\begin{table}[!ht]
\centering
\caption{
Amyloid-positivity decision reliability across representative MCSUVR
cut-offs using fixed 90\% conformal prediction intervals.
MRI-free point-estimate decisions are compared with the corresponding
real MRI-based reference decisions.
Captured disagreements denote discordant point-estimate decisions that were
designated indeterminate by the conformal decision rule.
Confident disagreement denotes discordant cases remaining among subjects
assigned a confident decision.
}
\label{tab:supp:mcsuvr_threshold}
\small
\setlength{\tabcolsep}{3.5pt}
\begin{tabular}{lcccccc}
\toprule
\textbf{Cut-off} &
\textbf{Ref. +/--} &
\textbf{FP/FN} &
\textbf{Confident} &
\textbf{Indet.} &
\textbf{Captured} &
\textbf{Conf. disag.} \\
\midrule
1.30 & 24/63 & 0/0 & 84/87 & 3/87 & --  & 0/84 \\
1.35 & 22/65 & 1/1 & 83/87 & 4/87 & 2/2 & 0/83 \\
1.40 & 21/66 & 1/0 & 86/87 & 1/87 & 1/1 & 0/86 \\
\textbf{1.42} & 21/66 & 0/0 & 85/87 & 2/87 & -- & 0/85 \\
1.45 & 20/67 & 0/0 & 85/87 & 2/87 & -- & 0/85 \\
1.50 & 20/67 & 0/0 & 86/87 & 1/87 & -- & 0/86 \\
\bottomrule
\end{tabular}
\end{table}

\FloatBarrier
\subsection{External transfer of conformal calibration to GARD}
\label{sec:supp:gard_conformal}

We further examined whether conformal calibration derived from OASIS-3
transferred to the independent GARD cohort. The same nonconformity score
used in the main analysis, defined as the absolute MRI-free quantification
error normalized by the subject-specific ensemble standard deviation, was
used throughout this analysis. For the conformal analysis, we included
82 GARD subjects with successful FreeSurfer processing and valid PET
quantification results for all ensemble members of the proposed model. For cross-cohort transfer, conformal
quantiles were estimated from the OASIS-3 calibration set and applied
directly to GARD without recalibration.

Direct transfer of the OASIS-3 calibration resulted in substantial
undercoverage on GARD (Table~\ref{tab:supp:gard_conformal}). For MCSUVR,
target coverage levels of 80\%, 90\%, and 95\% resulted in empirical
coverages of 42.7\%, 62.2\%, and 78.0\%, respectively. Undercoverage was
also observed across all six regional SUVR measures at the 90\% target
level, with empirical coverage ranging from 57.3\% in the precuneus to
86.6\% in the anterior cingulate cortex.

We therefore evaluated whether coverage could be recovered using
calibration data drawn from the GARD cohort. The 82 eligible GARD subjects were randomly divided into equal calibration
and test sets ($N=41$ each), and this split-half procedure was repeated
1,000 times to reduce dependence on a particular random partition. With GARD-specific calibration, empirical
MCSUVR coverage returned close to the requested levels, reaching
$81.4 \pm 8.8$\%, $90.7 \pm 6.2$\%, and $95.2 \pm 4.6$\% at target
coverage levels of 80\%, 90\%, and 95\%, respectively. At the 90\%
target level, mean regional coverage across the repeated splits ranged
from 90.2\% to 90.7\%. Prediction intervals were correspondingly wider
than those obtained by directly transferring the OASIS-3 calibration.
These results show that GARD-specific recalibration yielded empirical
coverage close to the target marginal coverage, whereas calibration
derived from OASIS-3 did not transfer directly to the external cohort.

\begin{table*}[!ht]
\centering
\caption{
External transfer and target-cohort recalibration of conformal prediction
on GARD using the ensemble-normalized nonconformity score.
For OASIS-3 calibration $\rightarrow$ GARD, conformal quantiles estimated
from the OASIS-3 calibration set were applied directly to the GARD
cohort.
For GARD-specific calibration, coverage is reported as mean $\pm$ SD across
1,000 random split-half partitions, and width denotes the corresponding
mean prediction interval width averaged across splits.
}
\label{tab:supp:gard_conformal}

\small
\setlength{\tabcolsep}{5pt}

\begin{tabular}{llcccc}
\toprule
\textbf{Measure}
& \textbf{Target}
& \multicolumn{2}{c}{\textbf{OASIS-3 cal. $\rightarrow$ GARD}}
& \multicolumn{2}{c}{\textbf{GARD-specific calibration}} \\
\cmidrule(lr){3-4}
\cmidrule(lr){5-6}
&
& \textbf{Coverage}
& \textbf{Width}
& \textbf{Coverage}
& \textbf{Width} \\
\midrule

\multicolumn{6}{l}{\textit{Composite MCSUVR}} \\

MCSUVR
& 80\%
& 42.7\%
& 0.075
& $81.4 \pm 8.8$\%
& 0.160 \\

& 90\%
& 62.2\%
& 0.119
& $90.7 \pm 6.2$\%
& 0.203 \\

& 95\%
& 78.0\%
& 0.151
& $95.2 \pm 4.6$\%
& 0.252 \\

\midrule

\multicolumn{6}{l}{\textit{Regional SUVR at 90\% target}} \\

Frontal
& 90\%
& 70.7\%
& 0.139
& $90.5 \pm 6.3$\%
& 0.208 \\

Parietal
& 90\%
& 65.9\%
& 0.108
& $90.6 \pm 6.4$\%
& 0.200 \\

Lateral temporal
& 90\%
& 69.5\%
& 0.098
& $90.7 \pm 6.1$\%
& 0.186 \\

Precuneus
& 90\%
& 57.3\%
& 0.141
& $90.6 \pm 6.4$\%
& 0.369 \\

Anterior cingulate
& 90\%
& 86.6\%
& 0.250
& $90.4 \pm 6.2$\%
& 0.280 \\

Posterior cingulate
& 90\%
& 79.3\%
& 0.184
& $90.2 \pm 6.4$\%
& 0.278 \\

\bottomrule
\end{tabular}
\end{table*}

\FloatBarrier  %

\enlargethispage{2\baselineskip}%
\begin{suppreferences}
\item\label{sref:choo2024slice} \bibinfo{author}{K.~Choo}, \bibinfo{author}{Y.~Jun}, \bibinfo{author}{M.~Yun}, \bibinfo{author}{S.~J. Hwang}, \bibinfo{title}{Slice-consistent 3d volumetric brain ct-to-mri translation with 2d brownian bridge diffusion model}, in: \bibinfo{booktitle}{International Conference on Medical Image Computing and Computer-Assisted Intervention}, \bibinfo{organization}{Springer}, \bibinfo{year}{2024}, pp. \bibinfo{pages}{657--667}.
\item\label{sref:kim2024adaptive} \bibinfo{author}{J.~Kim}, \bibinfo{author}{H.~Park}, \bibinfo{title}{Adaptive latent diffusion model for 3D medical image to image translation: Multi-modal magnetic resonance imaging study}, in: \bibinfo{booktitle}{WACV}, \bibinfo{year}{2024}, pp. \bibinfo{pages}{7604--7613}.
\item\label{sref:kim2025automated} \bibinfo{author}{D.~Kim}, \bibinfo{author}{K.~Choo}, \bibinfo{author}{S.~Lee}, \bibinfo{author}{S.~Kang}, \bibinfo{author}{M.~Yun}, \bibinfo{author}{J.~Yang}, \bibinfo{title}{Automated quantification of brain PET in PET/CT using deep learning-based CT-to-MR translation: a feasibility study}, \bibinfo{journal}{Eur.\ J.\ Nucl.\ Med.\ Mol.\ Imaging} \bibinfo{volume}{52} (\bibinfo{year}{2025}) \bibinfo{pages}{2959--2967}.
\item\label{sref:isola2017imagetoimage} \bibinfo{author}{P.~Isola}, \bibinfo{author}{J.-Y. Zhu}, \bibinfo{author}{T.~Zhou}, \bibinfo{author}{A.~A. Efros}, \bibinfo{title}{Image-to-image translation with conditional adversarial networks}, in: \bibinfo{booktitle}{CVPR}, \bibinfo{year}{2017}, pp. \bibinfo{pages}{1125--1134}.
\item\label{sref:mao2017least} \bibinfo{author}{X.~Mao}, \bibinfo{author}{Q.~Li}, \bibinfo{author}{H.~Xie}, \bibinfo{author}{R.~Y.~K. Lau}, \bibinfo{author}{Z.~Wang}, \bibinfo{author}{S.~P. Smolley}, \bibinfo{title}{Least squares generative adversarial networks}, in: \bibinfo{booktitle}{ICCV}, \bibinfo{year}{2017}, pp. \bibinfo{pages}{2794--2802}.
\item\label{sref:edison2013} \bibinfo{author}{P.~Edison}, \bibinfo{author}{S.~F. Carter}, \bibinfo{author}{J.~O. Rinne}, \bibinfo{author}{G.~Gelosa}, \bibinfo{author}{K.~Herholz}, \bibinfo{author}{A.~Nordberg}, \bibinfo{author}{D.~J. Brooks}, \bibinfo{author}{R.~Hinz}, \bibinfo{title}{Comparison of MRI based and PET template based approaches in the quantitative analysis of amyloid imaging with PIB-PET}, \bibinfo{journal}{NeuroImage} \bibinfo{volume}{70} (\bibinfo{year}{2013}) \bibinfo{pages}{423--433}.
\item\label{sref:landau2023quantification} \bibinfo{author}{S.~M. Landau}, \bibinfo{author}{T.~J. Ward}, \bibinfo{author}{A.~Murphy}, \bibinfo{author}{L.~Iaccarino}, \bibinfo{author}{T.~M. Harrison}, \bibinfo{author}{R.~La Joie}, \bibinfo{author}{S.~Baker}, \bibinfo{author}{R.~A. Koeppe}, \bibinfo{author}{W.~J. Jagust}, \bibinfo{author}{Alzheimer's Disease Neuroimaging Initiative}, \bibinfo{title}{Quantification of amyloid beta and tau PET without a structural MRI}, \bibinfo{journal}{Alzheimer's \& Dementia} \bibinfo{volume}{19} (\bibinfo{year}{2023}) \bibinfo{pages}{444--455}.
\item\label{sref:klunk2015centiloid} \bibinfo{author}{W.~E. Klunk}, \bibinfo{author}{R.~A. Koeppe}, \bibinfo{author}{J.~C. Price}, \bibinfo{author}{T.~L.~S. Benzinger}, \bibinfo{author}{M.~D. Devous}, \bibinfo{author}{W.~J. Jagust}, \bibinfo{author}{K.~A. Johnson}, \bibinfo{author}{C.~A. Mathis}, \bibinfo{author}{D.~Minhas}, \bibinfo{author}{M.~J. Pontecorvo}, \bibinfo{author}{C.~C. Rowe}, \bibinfo{author}{D.~M. Skovronsky}, \bibinfo{author}{M.~A. Mintun}, \bibinfo{title}{The Centiloid project: Standardizing quantitative amyloid plaque estimation by PET}, \bibinfo{journal}{Alzheimer's \& Dementia} \bibinfo{volume}{11} (\bibinfo{year}{2015}) \bibinfo{pages}{1--15.e4}.
\item\label{sref:ashburner2005unified} \bibinfo{author}{J.~Ashburner}, \bibinfo{author}{K.~J. Friston}, \bibinfo{title}{Unified segmentation}, \bibinfo{journal}{NeuroImage} \bibinfo{volume}{26} (\bibinfo{year}{2005}) \bibinfo{pages}{839--851}.
\item\label{sref:presotto2018ct} \bibinfo{author}{L.~Presotto}, \bibinfo{author}{L.~Iaccarino}, \bibinfo{author}{A.~Sala}, \bibinfo{author}{E.~G. Vanoli}, \bibinfo{author}{C.~Muscio}, \bibinfo{author}{A.~Nigri}, \bibinfo{author}{M.~G. Bruzzone}, \bibinfo{author}{F.~Tagliavini}, \bibinfo{author}{L.~Gianolli}, \bibinfo{author}{D.~Perani}, \bibinfo{author}{V.~Bettinardi}, \bibinfo{title}{Low-dose CT for the spatial normalization of PET images: A validation procedure for amyloid-PET semi-quantification}, \bibinfo{journal}{NeuroImage: Clinical} \bibinfo{volume}{20} (\bibinfo{year}{2018}) \bibinfo{pages}{153--160}.
\item\label{sref:matsuda2021amyloid} \bibinfo{author}{H.~Matsuda}, \bibinfo{author}{T.~Yamao}, \bibinfo{author}{M.~Shakado}, \bibinfo{author}{Y.~Shigemoto}, \bibinfo{author}{K.~Okita}, \bibinfo{author}{N.~Sato}, \bibinfo{title}{Amyloid PET quantification using low-dose CT-guided anatomic standardization}, \bibinfo{journal}{EJNMMI Research} \bibinfo{volume}{11} (\bibinfo{year}{2021}) \bibinfo{pages}{125}.
\end{suppreferences}

\end{document}